\documentclass[lettersize,journal]{IEEEtran}
\usepackage{amsmath,amsfonts}
\usepackage{algorithm}
\usepackage{algpseudocode}
\usepackage{hyperref}
\usepackage{array}
\usepackage{courier}
\usepackage{xcolor}
\hypersetup{hidelinks}
\usepackage{subfigure}
\usepackage{textcomp,soul}
\usepackage{stfloats}
\usepackage{url}
\usepackage{verbatim}
\usepackage{graphicx}
\usepackage{booktabs}
\usepackage{multirow}
\usepackage{cite}
\usepackage{amsthm}
\usepackage{siunitx}
\usepackage{bm}
\usepackage{pifont}
\theoremstyle{definition}

\makeatletter
\pretocmd\@bibitem{\color{black}\csname keycolor#1\endcsname}{}{\fail}
\newcommand\citecolor[1]{\@namedef{keycolor#1}{\color{blue}}}
\makeatother 
\title{VLM-UDMC: VLM-Enhanced Unified Decision-Making and Motion Control\\ for Urban Autonomous Driving}
\author{Haichao Liu, Haoren Guo, Pei Liu, Benshan Ma, Yuxiang Zhang, Jun Ma, and Tong Heng Lee
\thanks{Haichao Liu is with the Robotics and Autonomous Systems Thrust, The Hong Kong University of Science and Technology (Guangzhou), Guangzhou 511453, China, and also with the Department of Electrical and Computer Engineering, National University of Singapore, Singapore (e-mail: haichao.liu@u.nus.edu).}
\thanks{Haoren Guo, Yuxiang Zhang, and Tong Heng Lee are with the Department of Electrical and Computer Engineering, National University of Singapore, Singapore (e-mail: haorenguo\_06@u.nus.edu; zhangyx@nus.edu.sg; eleleeth@nus.edu.sg).}
\thanks{Pei Liu and Benshan Ma are with the Robotics and Autonomous Systems Thrust, The Hong Kong University of Science and Technology (Guangzhou), Guangzhou 511453, China (e-mail: pliu061@connect.hkust-gz.edu.cn, bma224@connect.hkust-gz.edu.cn)}
\thanks{Jun Ma is with the Robotics and Autonomous Systems Thrust, The Hong Kong University of Science and Technology (Guangzhou), Guangzhou 511453, China, and also with the Division of Emerging Interdisciplinary Areas, The Hong Kong University of Science and Technology, Hong Kong SAR, China (e-mail: jun.ma@ust.hk).} 
}
\begin{document}
\maketitle
\begin{abstract}
Scene understanding and risk-aware attentions are crucial for human drivers to make safe and effective driving decisions.
To imitate this cognitive ability in urban autonomous driving while ensuring the transparency and interpretability, we propose a vision-language model (VLM)-enhanced unified decision-making and motion control framework, named VLM-UDMC.
This framework incorporates scene reasoning and risk-aware insights into an upper-level slow system, which dynamically reconfigures the optimal motion planning for the downstream fast system. The reconfiguration is based on real-time environmental changes, which are encoded through context-aware potential functions.
More specifically, the upper-level slow system employs a two-step reasoning policy with Retrieval-Augmented Generation (RAG), leveraging foundation models to process multimodal inputs and retrieve contextual knowledge, thereby generating risk-aware insights.
Meanwhile, a lightweight multi-kernel decomposed LSTM provides real-time trajectory predictions for heterogeneous traffic participants by extracting smoother trend representations for short-horizon trajectory prediction.
The effectiveness of the proposed VLM-UDMC framework is verified via both simulations and real-world experiments with a full-size autonomous vehicle. It is demonstrated that the presented VLM-UDMC effectively leverages scene understanding and attention decomposition for rational driving decisions, thus improving the overall urban driving performance. Our open-source project is available at \href{https://github.com/henryhcliu/vlmudmc.git}{https://github.com/henryhcliu/vlmudmc.git}.
\end{abstract}

\section{Introduction}

\begin{figure}[t]
\centering
\includegraphics[width=0.95\linewidth]{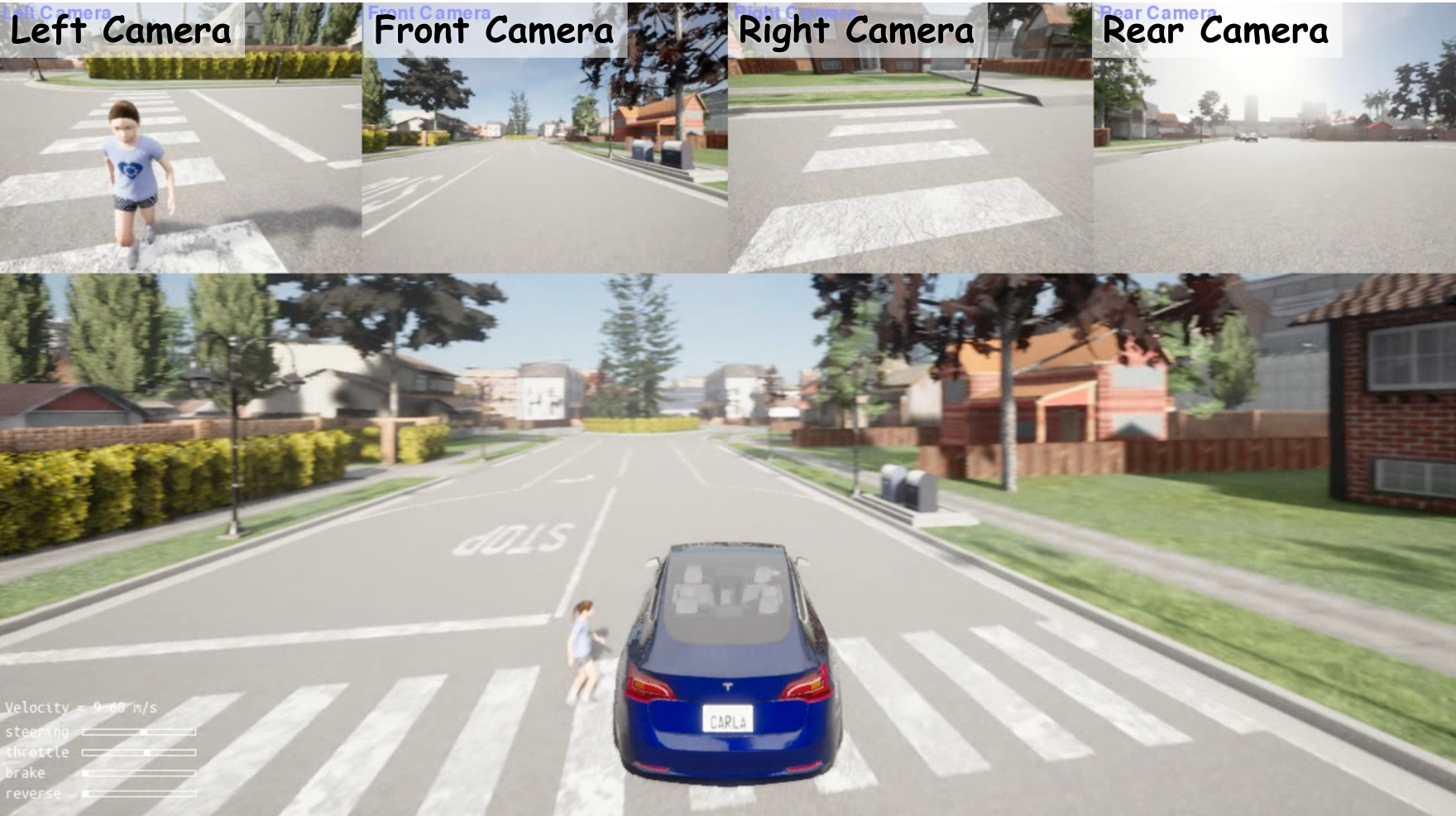}
\caption{The ego vehicle is turning left at an unsignalized T-intersection, while a little girl is crossing the road. The VLM deployed on the ego vehicle understands and inferences from the environment based on the real-time images captured by the onboard cameras, as shown from the four perspectives.}
\label{fig:overrallDemoCARLA}
\end{figure}

Urban autonomous driving has emerged as a critical technology to address the escalating challenges of traffic congestion, safety risks, and operational inefficiency in populated cities~\cite{chen2022milestones}. 
To tackle dynamic traffic scenarios involving diverse participants and long-tail events like the situation shown in Fig.~\ref{fig:overrallDemoCARLA}, future autonomous systems necessitate going beyond reactive control~\cite{gao2022optimal,zhang2023adaptive}, which requires advanced predictive and reasoning capabilities grounded in contextual understanding and explainable decision-making~\cite{gao2024enhance,zhao2024enhanced}.
Essentially, safe and efficient urban navigation requires machines to replicate human drivers' ability to integrate scene understanding, risk assessment, and adaptive decision-making, while ensuring transparency and interpretability for real-world safety-critical deployment~\cite{jia2023interactive}.
Recent progress in large language models (LLMs) and vision-language models (VLMs) has opened new avenues for designing such cognitive-driven driving systems. 
Unlike traditional modular or end-to-end frameworks that often lack interpretability and generalization, our approach leverages the representational power of the foundation models to enable complex scene understanding, adaptive planning, and transparent reasoning, key attributes for robust urban autonomy.

Existing approaches to autonomous driving can be broadly categorized into hierarchical frameworks~\cite{fan2018baidu} and end-to-end architectures~\cite{jiang2023vad}. Hierarchical methods decompose the driving task into modular layers, such as perception, prediction, planning, and control~\cite{chen2023milestones2}. Early hierarchical systems rely heavily on rule-based logic, encoding explicit traffic regulations and heuristic strategies into predefined decision trees~\cite{zhang2017finite,qi2022hierarchical} or finite state machines (FSM)~\cite{dosovitskiy2017carla}. Generally, lateral and longitudinal motion commands are generated by iLQR and PID, respectively~\cite{pan2024safe}. While interpretable, these systems struggle with the unbounded variability of urban scenes, where unstructured inputs (e.g., unexpected pedestrian movements, ambiguous road signs) exceed preprogrammed rules. For more flexible and optimal driving, optimization-based methods, such as Model Predictive Control (MPC)~\cite{xu2019design,brudigam2021stochastic}, advance this paradigm by formulating driving as an optimal control problem (OCP) over a receding horizon, considering vehicle dynamics, safety constraints, and so on. However, traditional MPC-based frameworks employ static weight parameters for constant objective functions, such as predefined-trajectory tracking, safety, comfort, and efficiency~\cite{liu2025udmc}, which hinders adaptability to changing conditions, like traffic jams in strange driving scenarios.

\begin{figure}
\centering
\includegraphics[width=1.0\linewidth]{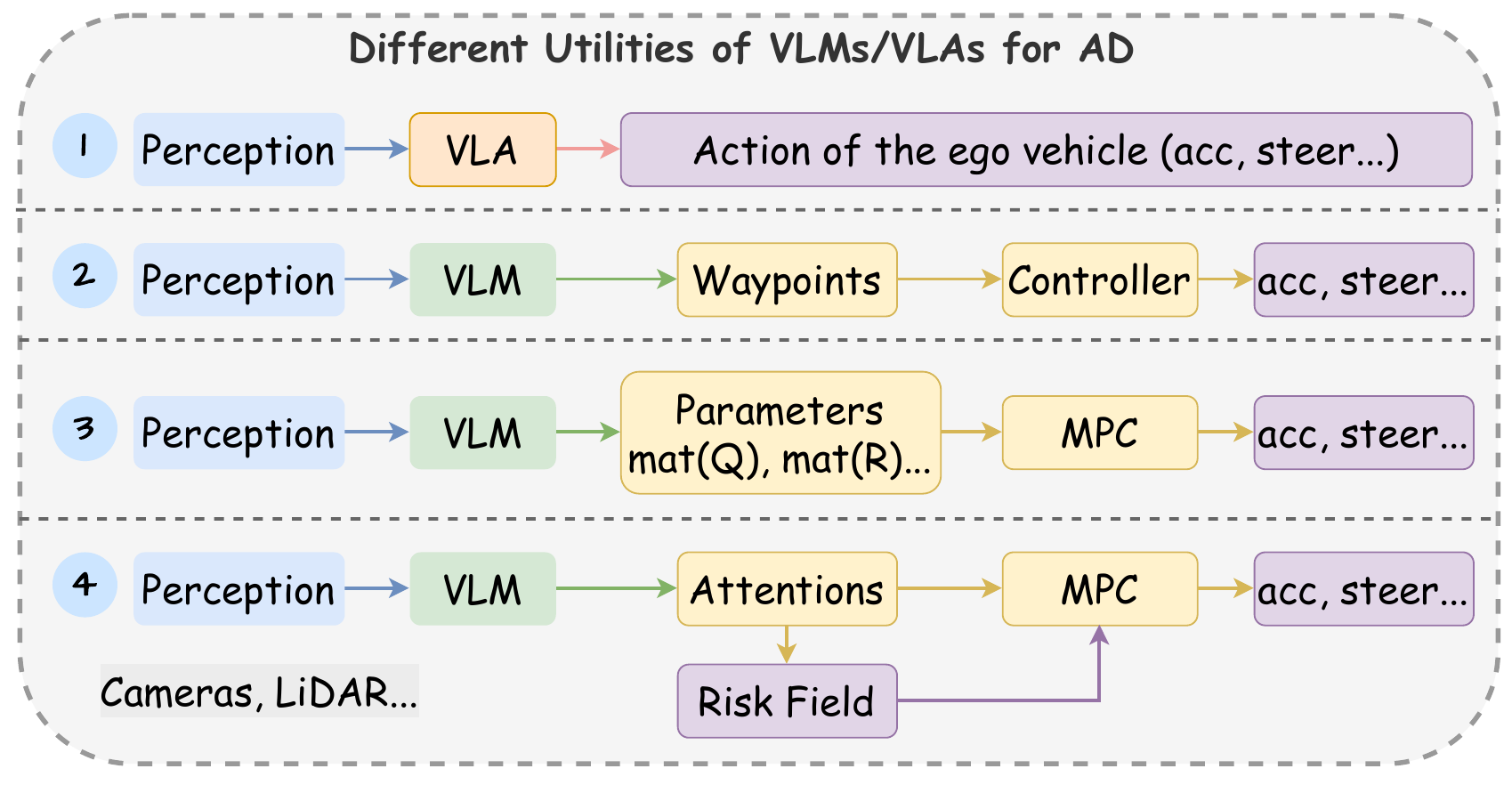}
\caption{The mainstream VLM/VLA utilities for autonomous driving. From top to bottom, the models are leveraged to generate the instant control commands, the predicted future waypoints to be followed by the low-level controller, the parameters to be integrated into MPC, and the risk attentions to be integrated into MPC, respectively.}
\label{fig:relatedWorksVLMVLA}
\end{figure}
On the other hand, end-to-end frameworks, such as UniAD~\cite{hu2023planning}, leverage Transformer architectures to model holistic driving policies, encoding sensor inputs directly into control commands or waypoints via self-attention and cross-attention mechanisms that capture implicit long-range interactions between agents~\cite{jiang2023vad, liu2025vlm,liu2025dsdrive}. In addition, Diffusion-based methods, such as DiffusionDrive~\cite{liao2025diffusiondrive}, enhance generative capabilities by offering multiple trajectory candidates for various future conditions involving surrounding traffic participants. With the emergence and development of LLMs, to synchronize driver interactions with the driving system, vision-language-action (VLA) approaches, like OpenDriveVLA~\cite{zhou2025opendrivevla}, are introduced to generate specific actions for the ego vehicle based on language input and real-time perception data, as illustrated in the first row of Fig.~\ref{fig:relatedWorksVLMVLA}. 
While these models exhibit strong performance in common scenarios during open-loop testing, they frequently experience diminished reliability in long-tail cases. Moreover, safety-critical decisions and actions require transparent reasoning rather than black-box inference~\cite{li2024think}, which is compromised by the direct action generation paradigm.
Consequently, as described in the last three rows of Fig.~\ref{fig:relatedWorksVLMVLA}, hybrid approaches that combine the semantic reasoning ability of VLMs with the structured optimization of model-based control are gaining more traction, aiming to balance flexibility with accountability~\cite{kuznietsov2024explainable,long2024vlm,atsuta2025lvlm,yao2024calmm}. 
However, existing hybrid approaches limit the potential of VLMs in semantic reasoning by only adaptively adjusting a few parameters within low-level driving modules, such as the weight matrices in the cost functions. Such minimal modifications are unlikely to bring about fundamental changes to the system's behavior, as the underlying optimization structure remains unchanged. Therefore, developing methods that enable adaptive reformulation of the optimization problem represents a promising research direction.

Another persistent challenge in existing hybrid methods is the limited integration of real-time motion forecasting and adaptive risk modeling~\cite{lstm_ts_ctcs, wen2023transformerstimeseriessurvey}. Many VLM-enhanced controllers, as shown in the third row of Fig.~\ref{fig:relatedWorksVLMVLA}, lack dynamic awareness of surrounding agents' future trajectories~\cite{tian2024drivevlm}, leading to suboptimal and conservative decisions for extreme driving safety in interactive scenarios, such as merging with high-speed mixed traffic or collaborative vehicles~\cite {li2024dcoma,liu2024improved,liu2025codrivevlm}. Additionally, model-based components, such as controllers with receding horizon manner, often assume that obstacles are static during the time window, failing to update with evolving scene contexts~\cite{lu2025empowering,liu2024incremental}. To fill the research gap, in the time-series forecasting studies, both Long Short-Term Memory (LSTM) and Transformer-based models~\cite{lstm_ts_ctcs, wen2023transformerstimeseriessurvey} have demonstrated strong capabilities in capturing temporal patterns and enhancing prediction accuracy. However, LSTM struggles to capture non-local or cross-period patterns such as periodicity and abrupt changes, while Transformer-based models often suffer from under-representation when trained on small datasets.
DLinear~\cite{zeng2023transformers} has been proposed as a lightweight and efficient alternative, particularly suitable for small datasets. By explicitly decomposing time series into trend and seasonal components, DLinear achieves improved interpretability and forecasting performance.
Nevertheless, DLinear exhibits excessive sensitivity in trend extraction when applied to relatively short sequences, often resulting in noisy or unsmooth trend estimations. This limitation hampers its performance in short-horizon prediction tasks in autonomous driving, where the underlying trend may be less pronounced. Therefore, incorporating a multi-kernel moving average mechanism to capture smoother and more robust trajectory trend representations is a worthwhile endeavor.

Building upon our previous research~\cite{liu2025udmc}, we propose a VLM-enhanced unified decision-making and motion control framework (VLM-UDMC) for urban autonomous driving, which is designed to effectively emulate the scene understanding and risk-aware attention mechanisms of human drivers.
It consists of an upper slow decision-making system that incorporates scene reasoning and risk-aware insights, and a downstream fast motion control system that utilizes the high-level decisions to adaptively reconfigure the OCP.
To support this framework, we incorporate the Retrieval-Augmented Generation (RAG) technique alongside a two-step reasoning strategy. This integration enables foundation models to process multimodal inputs and retrieve relevant contextual knowledge, leading to the generation of interpretable, risk-aware insights. These insights are subsequently utilized to construct the potential functions, which characterize traffic factors, and are embedded into the objective function of the OCP to guide motion planning dynamically.
Moreover, a compact yet powerful multi-scale decomposed LSTM enables efficient and accurate trajectory predictions for heterogeneous traffic participants, supporting proactive anticipation of interactive behaviors and behavior refinement over future time steps.
Importantly, an in-context learning mechanism continuously updates the VLM using newly encountered real-world scenarios, and this improves the robustness to long-tail cases while enhancing explainability by grounding decisions in verifiable perceptual and linguistic cues.
Consequently, the key contributions of this work are summarized as follows:
\begin{itemize}
    \item We propose VLM-UDMC for urban autonomous driving that integrates scene reasoning for risk-aware decision-making in the upper slow system, which is subsequently used to adaptively reconfigure the optimal motion planning in the downstream fast system.
    \item For enhanced scene understanding with multimodal prompts from VLMs, we develop a two-step reasoning mechanism combined with RAG-based in-context learning for attention-aware driving, while also addressing the long-tail problem through adaptation to newly encountered scenarios.
    \item By integrating a lightweight multi-kernel LSTM for real-time trajectory forecasting with attention-aware objective updating, the OCP is adaptively informed by heterogeneous time-scale inputs, enabling seamless and real-time decision-making and motion control under the semantic guidance of VLMs.
    \item The simulation and real-world experiments demonstrate the effectiveness of the proposed VLM-UDMC. A comprehensive comparison with baseline models highlights the superiority of our scheme, while an ablation study assesses the necessity of each functionality.
\end{itemize}

The rest of the paper is organized as follows. Section II presents the preliminaries of the presented urban driving algorithm. Section III illustrates the construction of the OCP, especially the potential functions for invoking by VLMs. In Section IV, the organization of the multimodal prompt with in-context learning for VLMs is elaborated. Next, Section V details the comparisons and ablations in both simulation and real-world experiments. Finally, Section VI concludes this work with a discussion.
\section{Preliminaries}
\subsection{Notation}
In this paper, vectors and matrices are represented by boldface letters, with lowercase letters for vectors and uppercase letters for matrices. Specifically, $\bm{x} \in \mathbb{R}^n$ denotes a vector and $\bm{X} \in \mathbb{R}^{n \times m}$ denotes a matrix. For a vector $\bm{x}$, the segment from its $i$-th to $j$-th element is denoted as $\bm{x}{[i:j]}$. In the case of matrices, the $i$-th row is represented as $[\bm{X}]_{i,\cdot}$ and the $j$-th column as $[\bm{X}]_{\cdot,j}$. The symbols $\mathbb{N}$, $\mathbb B$, and $\mathbb R_+$ refer to the sets of non-negative integers, boolean values, and positive real numbers, respectively. For ease of representation, a matrix composed of vectors $[\bm{s}^1, \bm{s}^2, \ldots, \bm{s}^n]^\top$ is simplified as $[\bm{s}^i]$, where $i \in {1, 2, \ldots, n}$. Finally, the notation $[x]_+$ is used to represent the function $\max\{x, 0\}$.

For scenario-specific notations, $[\, \cdot\,]^i_\tau$ is employed to specify the state vector of the $i$-th entity at time step $\tau$. It is crucial to recognize that each entity has its own distinct state vector. For example, a vehicle's state vector includes attributes such as position, velocity, and heading angle. Similarly, the state vector for a road lane center consists of position and orientation, whereas a traffic light object's state vector is defined by the stop line's position. When the state vector is expressed in the autonomous vehicle frame, we append a superscript $[\, \cdot\,]^\text{ev}$ to the notation. 

\subsection{Optimal Control Problem for Urban Driving}\label{subsec:2-3OCP}
Optimization-based methods are widely used in autonomous driving and robot navigation due to their explainability and consistent performance across diverse scenarios. 
Within these methods, the MPC balances computational tractability with adaptability to dynamic environments, making it ideal for urban scenarios where interactions with heterogeneous traffic participants and changing road conditions demand proactive and forward-looking planning. The OCP within this framework is formulated to minimize a cost function that encapsulates reference trajectory tracking accuracy, control effort, passenger comfort, and safety regulations.
Therefore, the cost function $J$ is structured to prioritize multiple competing objectives over a prediction horizon of length $N$, defined as:
\begin{align}
\label{basic_cost}
\begin{aligned}
J(\bm X, \bm U, \bm O) &= \sum_{\tau=1}^N \|\boldsymbol x_{\text{ref},\tau} - \boldsymbol x_\tau\|^2_{\bm{Q}} + \sum_{\tau=1}^N \|\boldsymbol u_{\tau}\|^2_{\bm{R}} \\
& + \sum_{\tau=2}^N \|\bm u_{\tau} - \bm u_{\tau-1}\|^2_{\mathbf{R}_d} + F(\bm O, \bm X).
\end{aligned}
\end{align}
Here, $\bm X = \{\bm x_1, \bm x_2, \ldots, \bm x_N\} \in \mathbb{R}^{m \times N}$ represents the sequence of vehicle states over the horizon, with each $\bm x_\tau \in \mathbb{R}^m$ encoding the state of the autonomous vehicle at time step $\tau$. The control input sequence $\bm U = \{\bm u_1, \bm u_2, \ldots, \bm u_N\} \in \mathbb{R}^{q \times N}$ consists of the control input $u_i\in \mathbb R^q$ at each time step $\tau$. The reference trajectory $\bm x_{\text{ref},\tau}$ is dynamically derived according to the global path and the current vehicle state, ensuring alignment with high-level route plans.

Subsequently, the first term in $J$ penalizes deviations from the reference trajectory using a positive semi-definite diagonal weighting matrix $\bm{Q} \in \mathbb{R}^{m \times m}$, emphasizing accurate path following. The second term discourages excessive control effort through $\bm{R} \in \mathbb{R}^{q \times q}$, promoting energy-efficient maneuvers. The third term, weighted by $\mathbf{R}_d \in \mathbb{R}^{q \times q}$, minimizes abrupt changes in control inputs to enhance passenger comfort and reduce wear on vehicle actuators. The final term $F(\bm O, \bm X)$ integrates a risk field that encodes safety-critical interactions with surrounding objects (e.g., other vehicles, pedestrians, curbs) in $\bm O=\{o_1,o_2,\cdots, o_k\}$ by converting their states into potential forces, dynamically adjusting the cost landscape to avoid collisions.

In addition, respecting the physical dynamics and operational limits of the autonomous vehicle, the complete OCP is posed as a constrained optimization problem:
\begin{equation}
\label{NMPCOptProb}
\begin{array}{ll}
\underset{\{\boldsymbol{x_\tau}, \boldsymbol{u_\tau}\}}{\min} & J(\bm X, \bm U, \bm O) \\
\text{subject to} & \boldsymbol x_{\tau+1} = f(\boldsymbol x_\tau, \boldsymbol u_\tau), \quad \forall \tau \in \{1, 2, \ldots, N\} \\
& -\boldsymbol u_\text{min} \preceq \boldsymbol u_\tau \preceq \boldsymbol u_\text{max}, \quad \forall \tau \in \{1, 2, \ldots, N\} \\
& -\boldsymbol x_\text{min} \preceq \boldsymbol x_\tau \preceq \boldsymbol x_\text{max}, \quad \forall \tau \in \{1, 2, \ldots, N\},
\end{array}
\end{equation}
where the first constraint enforces the nonlinear vehicle dynamics model, which maps the current state $\boldsymbol x_\tau$ and control input $\boldsymbol u_\tau$ to the next state $\boldsymbol x_{\tau+1}$ (detailed in Section \ref{subsec:vehicle_dynamics}). The subsequent constraints bound the control inputs and states within feasible ranges dictated by the vehicle's physical limitations (e.g., maximum steering angle, acceleration capacity), ensuring safe and realizable maneuvers. By solving this problem iteratively at each time step with the latest sensor observations and reasoning decisions of the VLM (elaborated in Section~\ref{subsec:context_generation}), the framework adapts to evolving scenes while maintaining real-time performance, forming the basis for risk-aware control in urban environments.
\begin{figure*}
\centering
\includegraphics[width=0.8\linewidth]{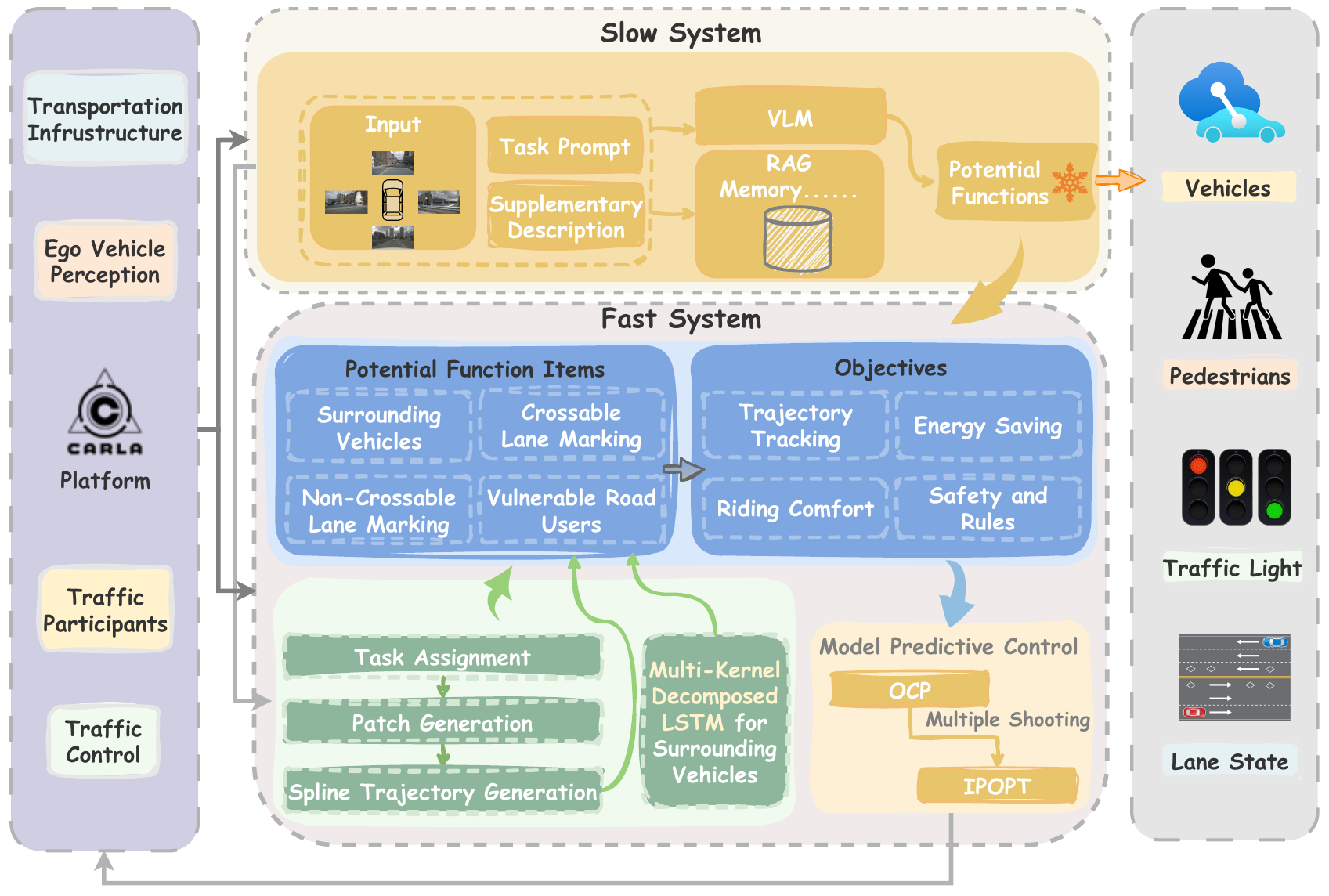}
\caption{Framework of the proposed VLM-enhanced risk-aware urban autonomous driving system is designed to address the complexities of urban driving. This framework comprises four sequential steps and incorporates an asynchronous mechanism to reconcile the latency inherent in scene understanding inference with the stringent demands of real-time control.}
\label{fig:drivevm_pipeline}
\end{figure*}
\subsection{Dynamics Model for Autonomous Vehicle}\label{subsec:vehicle_dynamics} 
In this work, we adopt a nonlinear vehicle dynamics model from~\cite{ge2021numerically}, as follows:
\begin{equation}
\label{eq:bicycle_dynamics}
\bm{x}_{\tau+1} = 
\left[\begin{array}{c}
p_{x,\tau} + T_s \left(v_{x,\tau} \cos \varphi_{\tau} - v_{y,\tau} \sin \varphi_{\tau}\right) \\
p_{y,\tau} + T_s \left(v_{y,\tau} \cos \varphi_{\tau} + v_{x,\tau} \sin \varphi_{\tau}\right) \\
\varphi_{\tau} + T_s \omega_{\tau} \\
v_{x,\tau} + T_s a_{\tau} \\
\frac{m_a v_{x,\tau} v_{y,\tau} + T_s L_k \omega_{\tau} - T_s k_f \delta_{\tau} v_{x,\tau} - T_s m_a v_{x,\tau}^2 \omega_{\tau}}{m_a v_{x,\tau} - T_s (k_f + k_r)} \\
\frac{I_z v_{x,\tau} \omega_{\tau} + T_s L_k v_{y,\tau} - T_s l_f k_f \delta_{\tau} v_{x,\tau}}{I_z v_{x,\tau} - T_s (l_f^2 k_f + l_r^2 k_r)}
\end{array}\right],
\end{equation}
which is chosen for its robust numerical stability and high-fidelity representation of complex motions. In (\ref{eq:bicycle_dynamics}), the state vector $\bm{x} = [p_x, p_y, \varphi, v_x, v_y, \omega]^\top \in \mathbb{R}^{6}$ captures six degrees of freedom: longitudinal and lateral positions, heading angle, longitudinal and lateral velocities, and yaw rate, respectively. The control input vector $\bm{u} = [a, \delta]^\top \in \mathbb{R}^{2}$ consists of longitudinal acceleration $a$ and front-wheel steering angle $\delta$, which act as the primary actuation commands.
Besides, the discrete-time dynamics (\ref{eq:bicycle_dynamics}), also expressed by $\bm{x}_{\tau+1} = f(\bm{x}_\tau, \bm{u}_\tau)$, is derived using the backward Euler method with a fixed time step $T_s=0.05$\,s.
Here, $m_a$ denotes the vehicle mass, $l_f$ and $l_r$ are the distances from the center of mass to the front and rear axles, respectively, while $k_f$ and $k_r$ represent the front and rear tire cornering stiffness, a key parameter for modeling lateral forces during steering. $I_z$ is the polar moment of inertia about the vertical axis, describing the vehicle’s resistance to yaw acceleration. The term $L_k = l_f k_f - l_r k_r$ simplifies the formulation by capturing the differential effect of front and rear tire forces on yaw dynamics.

\section{Spatial-Temporal Motion Planning and Control for Urban Driving}\label{sec:methodology}
This section presents and analyzes the mathematical modeling of traffic elements, including surrounding vehicles, vulnerable road users (VRUs), and components related to traffic rules, for invocation by VLMs, as illustrated in Fig.~\ref{fig:drivevm_pipeline}. Additionally, to enable spatial-temporal motion planning, we introduce a compact yet powerful trajectory prediction model based on LSTM networks and analyze the mechanism for capturing detailed historical data.

\subsection{Traffic Items Description and Modeling}\label{subsec:3-1apfModeling}
In urban autonomous driving, effective modeling of dynamic and static traffic elements is crucial for safe yet efficient trajectory planning. We employ potential field methods to represent four key categories of traffic items: nearby vehicles with collision risks, pedestrians as the primary VRUs, adjacent road lane markings for behavioral guidance, and recognized red traffic lights. These potential functions are dynamically activated based on scene understanding and inference from VLMs, ensuring that only relevant environmental factors with their specific ID $i,j,k,q\in \mathcal{S}_\tau$ influence the OCP at each time step. For example, when the VLM identifies that certain surrounding vehicles pose no collision risk (e.g., traveling in opposite directions on separate lanes), their corresponding potential functions are excluded from the OCP formulation, enhancing computational efficiency and decision relevance.

The potential field $F(\bm{p}_\tau^\text{env}, \bm{x}_\tau)$ at time step $\tau$ integrated into the OCP is a composite of multiple specialized potential functions, each designed to model specific traffic interactions:
\begin{equation}
\label{eq:sum_potential_fields}
\begin{aligned}
F(\bm{p}_\tau^\text{env}, \bm{x}_\tau) = &\sum_{i\in \mathcal{S}_\tau} F_{\text{NR},\tau}^i + \sum_{j\in \mathcal{S}_\tau} F_{\text{CR},\tau}^j+ \sum_{k\in \mathcal{S}_\tau} F_{\text{V},\tau}^k  \\
&+ \sum_{q\in \mathcal{S}_\tau} F_{\text{VRU},\tau}^q + F_{\text{TL},\tau},
\end{aligned}
\end{equation}
where $i$, $j$, $k$, and $q$ index non-crossable lane markings, crossable lane markings, surrounding risk vehicles, and VRUs, respectively. The vector $\bm{p}_\tau^\text{env}$ encapsulates the state information of all relevant traffic entities, including their positions, velocities, and orientations in the global Cartesian coordinate frame.

\subsubsection{Lane Marking Potential Functions}
The lateral distance between the autonomous vehicle and lane markings is a critical factor in defining repulsive potential forces. We represent the road centerline state vector as $\bm{p}^i = [p_x^i, p_y^i, \varphi^i]^\top$, where $p_x^i$ and $p_y^i$ denote the global coordinates of the centerline, and $\varphi^i$ is the tangential angle relative to the global $x$-axis. 
Consequently, using the plane geometric relationship, the lateral distance from the autonomous vehicle to a lane marking is calculated as follows:
\begin{equation}
s_{\text{R}}(\bm{x}, \bm{p}^i) = \sigma \left( \begin{bmatrix} \sin(\varphi^i) & \cos(\varphi^i) \end{bmatrix} \begin{bmatrix} p_x - p_x^i \\ p_y - p_y^i \end{bmatrix} \right) + \frac{w_R}{2},
\end{equation}
where \(\sigma\) is the indicator function denoting the direction of a specific lane marking: \(\sigma = 1\) if the lane marking is on the left side, and \(\sigma = -1\) if it is on the right. Additionally, \(w_R\) is a constant representing the lane width in the current traffic system.

\textbf{Crossable Lane Marking Items:}
For permissible lane changes, such as overtaking, the crossable lane marking potential \( F_{\text{CR}} \) guides the autonomous vehicle to perform lane-changing maneuvers and return to the lane center both before and after the maneuver. This potential function also assists the vehicle in maintaining its position in the middle of the road. It is defined as follows:
\begin{align}\label{eq:F_CR}
F_{\text{CR}}^j = 
\begin{cases} 
a_\text{CR} (s_{\text{R}}(\bm{x}, \bm{p}^j) - b_\text{CR})^2 & s_{\text{R}_j} < b_\text{CR} \\
0 & s_{\text{R}_j} \geq b_\text{CR}
\end{cases}
\end{align}
where \( a_\text{CR} \) determines the force intensity, and \( b_\text{CR} \) defines the effective range. The conditions in (\ref{eq:F_CR}) imply that if the autonomous vehicle remains within the lane without significant deviation, no force is exerted by the lane markings, ensuring smooth and stable driving control.

\textbf{Non-Crossable Lane Marking Items:}
To enforce traffic rules and prevent unsafe lane crossings (e.g., near curbs or sidewalks), we design a repulsive potential function $F_{\text{NR}}$ that increases orderly as the autonomous vehicle approaches the marking:
\begin{align}\label{eq:F_NR}
F_{\text{NR}}^i = 
\begin{cases} 
m_s & s_{\text{R}_i} \leq 0.1 \\ 
\frac{a_\text{NR}}{s_{\text{R}}(\bm{x}, \bm{p}^i)^{b_\text{NR}}} - e_s & 0.1 < s_{\text{R}_i} < 1.5 \\
0 & s_{\text{R}_i} \geq 1.5
\end{cases}
\end{align}
where $a_\text{NR}$ and $b_\text{NR}$ control the force intensity and effective range, and smooth transition parameters $e_s$ and $m_s$ ensure continuity by setting as
$$e_s = \frac{a_\text{NR}}{1.5^{b_\text{NR}}}, \quad m_s = \frac{a_\text{NR}}{0.1^{b_\text{NR}}} - e_s.$$

\subsubsection{Surrounding Vehicle Potential Functions}
We consider the geometric approximation for vehicle modeling as ellipsoid wrappings, designed to generate repulsive forces for collision avoidance.
For computational efficiency, we use an ellipsoid approximation to represent the surrounding vehicles with attention as follows:
\begin{equation}
F_{\text{V}}^k = \sum_{m=1}^2 \frac{a_\text{V} (r_a r_b)^2}{(r_b^2 (p_x^m - p_x^k)^2 + r_a^2 (p_y^m - p_y^k)^2)^{b_\text{V}}},
\end{equation}
where $r_a$ and $r_b$ are the ellipsoid major and minor semi-axes, respectively.

\subsubsection{Vulnerable Road User Potential Functions}
Pedestrians are modeled as a single circular potential centered at their head position in bird's-eye view:
\begin{equation}
F_{\text{VRU}}^q = \frac{a_\text{VRU}}{((p_x - p_x^q)^2 + (p_y - p_y^q)^2)^{b_\text{VRU}}},
\end{equation}
where $a_\text{VRU}$ and $b_\text{VRU}$ control the repulsive force intensity for the $q$-th pedestrian.

\subsubsection{Traffic Light Potential Function}
For red traffic lights, we design a potential that guides the autonomous vehicle to stop smoothly at the stop line while maintaining lateral control:
\begin{equation}
F_\text{TL} = \beta \left(\frac{a_{\text{TL}_1}}{d_x^\text{ev}} + \frac{a_{\text{TL}_2}}{d_{y,l}^\text{ev}} + \frac{a_{\text{TL}_2}}{d_{y,r}^\text{ev}}\right),
\end{equation}
where $\beta$ is a binary indicator: $\beta=1$ if the traffic light state is red, and $\beta=0$ vice versa. Besides, $a_{\text{TL}_1}$ and $a_{\text{TL}_2}$ are longitudinal and lateral intensity parameters, and $d_x^\text{ev}$, $d_{y,l}^\text{ev}$, $d_{y,r}^\text{ev}$ are the autonomous vehicle's distances to the stop line and lane markings, respectively.

The above potential field-based modeling structure, dynamically informed by VLM scene understanding, enables the VLM-UDMC framework to prioritize safety-critical interactions while maintaining computational efficiency in complex urban environments. Once the potential functions are integrated into the OCP, the framework also requires the predicted states of selected traffic elements to facilitate a receding horizon approach for urban driving.

\subsection{Real-Time Motion Prediction of Traffic Participants}
To accurately forecast the motion of heterogeneous traffic participants in urban scenarios, we propose an efficient prediction approach, as illustrated in Fig.~\ref{fig:smollstm}. It is a lightweight yet powerful model that integrates a multi-kernel moving average mechanism to extract smoother trend representations for short-horizon trajectory prediction. The subsequent LSTM module further captures both global and local temporal patterns, enabling more efficient, accurate, and robust forecasting of real-time motion prediction.

\subsubsection{Multiscale Decomposition for Trajectory Data} 
\begin{figure}[t]
\centering
\includegraphics[width=1\linewidth]{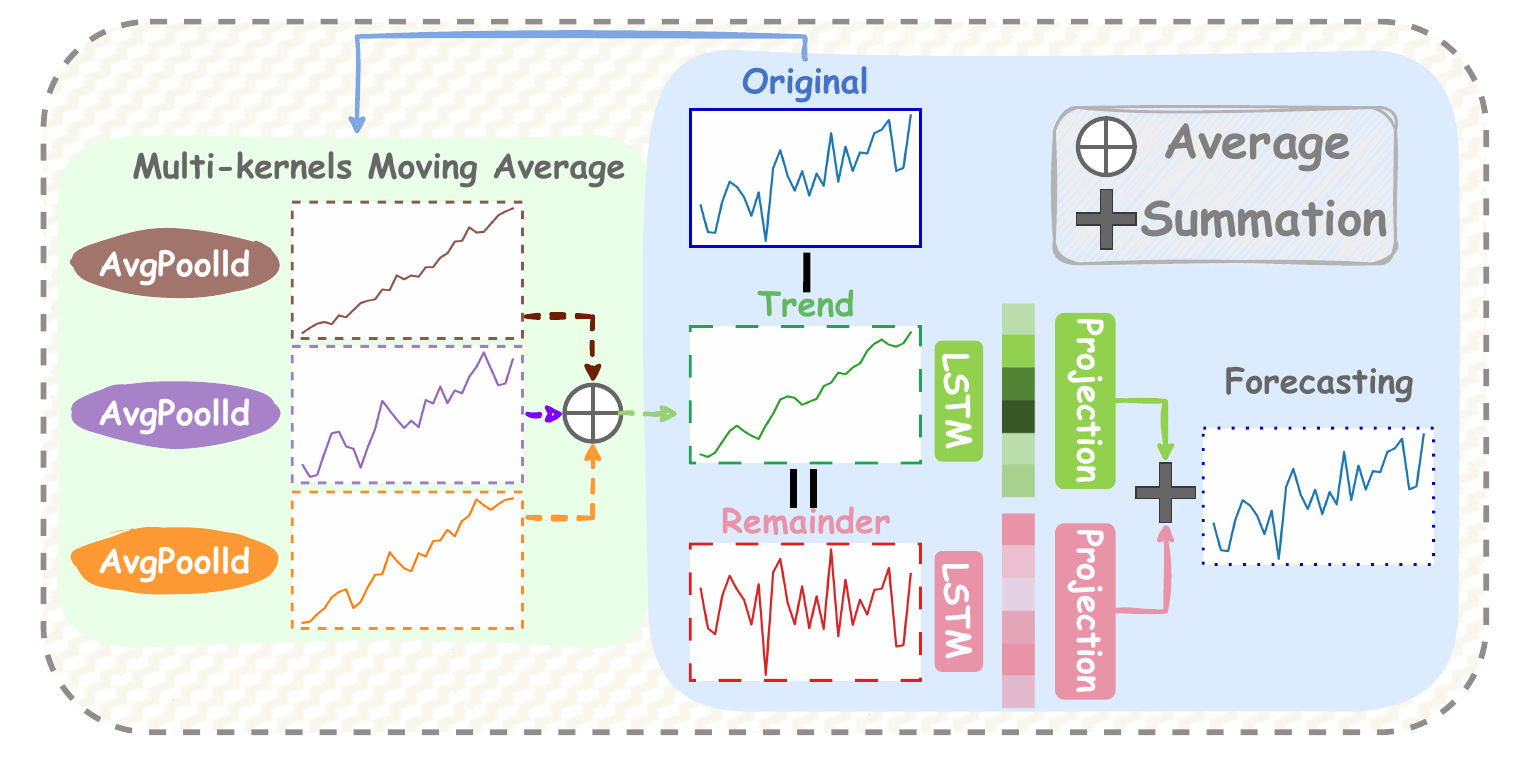}
\caption{The architecture of the proposed multi-kernel prediction approach for real-time motion prediction of the traffic participants, including surrounding vehicles and VRUs. It leverages the DLinear mechanism to smooth sparse trajectory data, improving the model's sensitivity to trend changes.}
\label{fig:smollstm}
\end{figure}

To generate time-series trajectory predictions from input data with a sparser time point distribution, we introduce a multi-kernels moving average mechanism based on 1D average pooling (\texttt{AvgPool1d}) for multiscale analysis. As shown in Fig.~\ref{fig:smollstm}, multiple \texttt{AvgPool1d} operations with different kernel sizes are applied to the original trajectory time series. This multiscale processing generates several smoothed representations of the input.

Mathematically, let the original trajectory time series be denoted as $\bm{X} = [x_1, x_2, \ldots, x_T]^\top$, where $T$ is the number of time points. For an \texttt{AvgPool1d} operation with kernel size $k$, the output $\bm{X}_{pool}^k$ at position $i$ is calculated as:
\begin{equation}
\bm{X}_{pool}^k[i]=\frac{1}{k}\sum_{j = i}^{i + k - 1}x_j.
\end{equation}
By using multiple kernel sizes (different $k$ values), we obtain multiple multiscale smoothed sequences. These sequences are then combined (via the average operation as indicated in the structure) to extract a more robust trend component.

The key advantage of this multiscale analysis is producing a smoother trend from the relatively sparse trajectory data. This smoother trend enhances the model's sensitivity to trend changes, resulting in the general movement patterns of traffic participants being more accurately.

\subsubsection{Trend and Remainder Component Separation}
After multiscale processing, we decompose the trajectory data into two components: trend ($\bm{T}$) and remainder ($\bm{R}$). The trend can be regarded as intra information, which is more general and global, representing the overall direction and pattern of the traffic participants' movement. In contrast, the remainder contains inter information, including more detailed and local features such as sudden speed changes or small-scale maneuvers.
The separation process can be described as:
\begin{equation}
\bm{X}=\bm{T}+\bm{R},
\end{equation}
where $\bm{T}$ is obtained from the multiscale-smoothed and combined sequence, and $\bm{R}$ is the difference between the original sequence and the trend sequence.

\subsubsection{LSTM-Based Prediction for Components}
Each of these components (trend and remainder) is then fed into separate LSTM networks. The LSTM is well-suited for processing sequential data and can capture the temporal dependencies within the trend and remainder components. After passing through the LSTM layers, projection layers are used to transform the LSTM outputs into the appropriate prediction space.

For the trend component, let the LSTM for trend be $\text{LSTM}_T$ and the projection layer be $\text{Proj}_T$. The prediction for the trend component $\hat{\bm{T}}$ is:
\begin{equation}
\hat{\bm{T}}=\text{Proj}_T(\text{LSTM}_T(\bm{T})).
\end{equation}
Similarly, for the remainder component, with $\text{LSTM}_R$ and projection layer $\text{Proj}_R$, the prediction $\hat{\bm{R}}$ is:
\begin{equation}
\hat{\bm{R}}=\text{Proj}_R(\text{LSTM}_R(\bm{R})).
\end{equation}
Finally, the overall motion prediction $\hat{\bm{X}}$ is obtained by summing the predictions of the two components:
\begin{equation}
\hat{\bm{X}}=\hat{\bm{T}}+\hat{\bm{R}}.
\end{equation}

This two-component prediction approach, enabled by multiscale decomposition, not only improves the model's ability to capture the overall movement trend of traffic participants but also preserves key local details. As a result, it significantly enhances the accuracy of real-time motion prediction in complex urban traffic environments, providing reliable information for the subsequent planning and control of autonomous vehicles. 
\section{Multimodal Scene Understanding and Driving Attention Adaptation}
This section addresses the fast-slow mechanism designed to reconcile the slow inference time of VLMs with the rapid control demands of the OCP. It also explores multimodal prompt context generation and RAG-based in-context learning for the urban driving system.
\subsection{Time-Scale Architecture of the Urban Driving System}\label{subsec:fast_slow}
In urban autonomous driving, the need to balance real-time responsiveness and semantic reasoning efficiency gives rise to a fast-slow architecture. Our specific design addresses two key challenges: the disparity in computational demands between low-level control and high-level scene understanding, and the varying relevance of semantic updates based on evolving driving context.

\subsubsection{Motivation and Necessity}
Semantic attention and MPC-based motion planning are updated on different time scales to ensure both improved system performance and efficiency. The low-level MPC, as defined by (\ref{NMPCOptProb}), requires high-frequency actuation (20\,Hz in our system) to react to dynamic urban environments, such as sudden pedestrian movements or vehicle cut-ins. In contrast, VLM inference for scene understanding incurs higher latency, with inference times ranging from 0.3 to 1.5\,s for the foundation models like SmolVLM~\cite{marafioti2025smolvlm} and Qwen3-4B~\cite{qwen3technicalreport}. 
 
The diversity of driving scenarios also motivates the design of a hierarchical framework with updates operating at different time scales. For instance, on a highway with a stable lane-keeping task, attention to road boundaries and forward traffic suffices, and frequent VLM re-inference probably adds no value. Conversely, considering dynamic scenarios (e.g., navigating a complex intersection), the objects for attention must adapt quickly (e.g., crossing pedestrians, changing traffic lights). Thus, the risk zone mechanism is proposed to identify critical scenarios. 

The proposed fast-slow architecture decouples as follows: \textbf{Fast sub-system} with high-frequency MPC with 20\,Hz for control, using a static OCP formulation between semantic updates.
\textbf{Slow sub-system} with low-frequency foundation model-driven attention adaptation with $0.3-1.5$\,Hz, refining the OCP based on scene semantics.

\subsubsection{Mathematical Formulation of the Fast-Slow Mechanism}
Let $t_k$ denote discrete time steps, with $k \in \mathbb{N}$. The fast sub-system, MPC, operates at a high frequency, solving the OCP at each $t_k$:
\begin{equation}
\begin{aligned}
& \underset{\{\boldsymbol{x}_\tau, \boldsymbol{u}_\tau\}_{\tau=k}^{k+N-1}}{\text{min}}
& & J(\boldsymbol{X}, \boldsymbol{U}, \boldsymbol{P}(t_m)) \\
& \text{subject to}
& & \boldsymbol{x}_{\tau+1} = f(\boldsymbol{x}_\tau, \boldsymbol{u}_\tau), \\
& & & \boldsymbol{u}_\text{min} \preceq \boldsymbol{u}_\tau \preceq \boldsymbol{u}_\text{max}, \\
& & & \boldsymbol{x}_\text{min} \preceq \boldsymbol{x}_\tau \preceq \boldsymbol{x}_\text{max},
\end{aligned}
\end{equation}
where $t_m \leq t_k$ is the most recent time step when the slow sub-system with a VLM updated the OCP items $\boldsymbol{P}$ to formulate $F(\bm{p}_{t_k}^\text{env})$ in (\ref{eq:sum_potential_fields}).
In addition, to enhance the inference stability and regulate the final decision format, before updating $\boldsymbol{P}$, the slow sub-system retrieves relevant memory items $\mathcal{M} = \{m_1, m_2, \ldots, m_M\}$, which contains similar scene embeddings and the relevant responses, and generating a new OCP formulation via foundation model inference:
\begin{equation}
\boldsymbol{P}(t_m) = \text{FM}\left(C(t_m),\text{Retrieve}(\mathcal{M}, \mathcal{C}(t_m))\right),
\end{equation}
where $\mathcal{C}(t_m)$ is the multimodal context including the RGB images from the cameras mounted on the autonomous vehicle and the supplementary text describing the task and the response format of the VLM. More detailed descriptions about the prompt can be found in Section~\ref{subsec:context_generation}. Between updates at a time step $t_k$, the fast sub-system reuses the insights from $\boldsymbol{P}(t_m)$ to solve the OCP, ensuring continuity:
\begin{equation}
\boldsymbol{P}(t_k) = \boldsymbol{P}(t_m), t_k \in \{t\,|\,t_m <t< t_{m+1},t\in\mathbb R_+\}.
\end{equation}

\subsubsection{Adaptive OCP Evolution}
The key insight is that the OCP’s cost function and constraints evolve only when the slow sub-system provides a meaningful change in scene semantics. For instance, if the VLM infers a new pedestrian entering the autonomous vehicle’s vicinity and the right road lane changes from crossable to non-crossable, it updates the potential field term $F(\boldsymbol{P}, \boldsymbol{X})$ in the cost function by
\begin{equation}
F(\boldsymbol{P}(t_{m+1}), \boldsymbol{X}) = F(\boldsymbol{P}(t_m), \boldsymbol{X}) + F_{\text{VRU}}^1-F_\text{CR}^2+F_\text{NR}^2,
\end{equation}
where $F_{\text{VRU}}^1$ is a new repulsive potential for the pedestrian with ID$=1$, $F_\text{NR}^2$ is the potential force from the non-crossable right road lane, while its adversary $F_\text{CR}^2$ is removed from the potential field. Nevertheless, if no new objects or semantic changes are detected, $\boldsymbol{P}(t_{m+1}) = \boldsymbol{P}(t_m)$, and the fast sub-system continues with the existing OCP.

This architecture ensures that the autonomous vehicle reacts to immediate threats via high-frequency MPC while adapting to long-term scene changes through low-frequency VLM inference. By decoupling temporal scales, it balances computational efficiency and semantic richness, enabling safe and responsive urban driving.
\begin{figure}[t]
    \centering
    \includegraphics[width=1\linewidth]{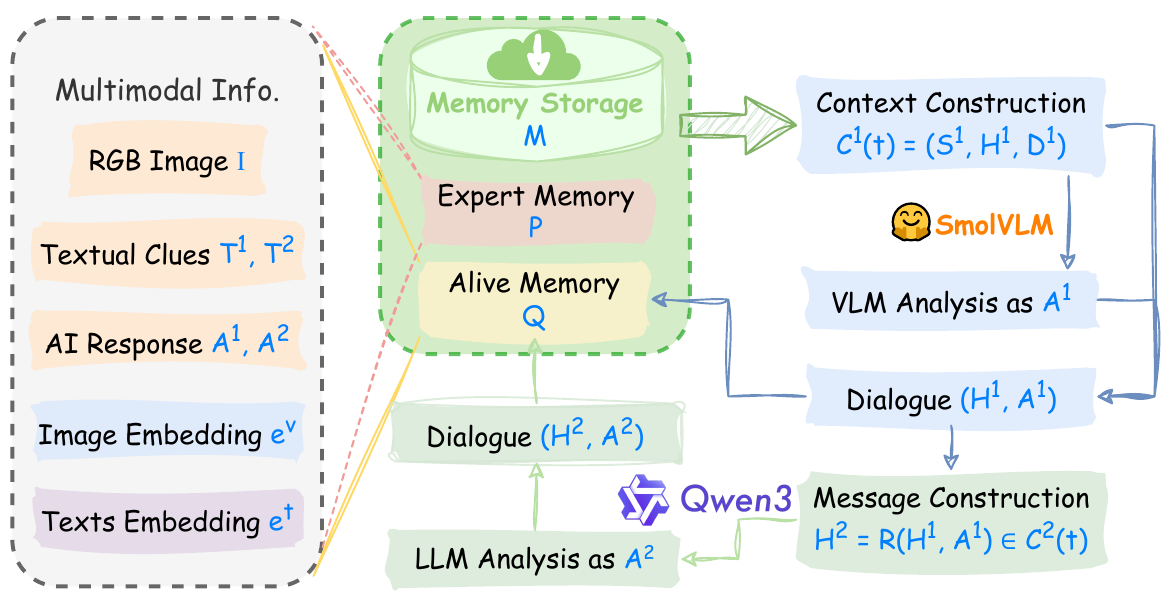}
    \caption{Pipeline of the prompt generation and RAG process for the proposed scene understanding and risk assessment mechanism.}
    \label{fig:memory_RAG}
\end{figure}
\subsection{Multimodal Context Generation for Scene Understanding}\label{subsec:context_generation}
To enable foundation models to reason about complex urban driving scenarios, as illustrated in Fig.~\ref{fig:memory_RAG}, we design a two-step structured multimodal context that integrates visual inputs, textual information, and historical knowledge with a VLM and an LLM. The corresponding prompts for the first and second steps are denoted by $\mathcal{C}^1$ and $\mathcal{C}^2$, respectively. The elements of the prompts also have their superscripts to indicate the step order. Inspired by~\cite{li2025drive}, this context serves as the foundation to take advantage of the VLM's image understanding ability and LLM's specific reasoning ability for driving attention adaptation, guiding the dynamic reformulation of the OCP. Concretely, the visual information is analyzed by a VLM to generate a structured textual description of the images. Then, the second step is to get the final decision of attention from an LLM. Note that the prompt to the LLM is generated by reformulating the response of VLM and adding necessary supplementary descriptions.

\subsubsection{Context Composition Framework}

The prompt $\mathcal{C}$ is constructed as a tuple of three primary components:
\begin{equation}
\mathcal{C} = (\mathcal{S}, \mathcal{H}, \mathcal{D}),
\end{equation}
where $\mathcal{S}$ represents the system message, defining the foundation model's task and response format, $\mathcal{H}$ denotes the human message, containing real-time visual observations (for VLM) and supplementary spatial information, $\mathcal{D}$ encapsulates analogical dialogue examples retrieved from memory storage for in-context learning.

\subsubsection{System Message Structure}
\begin{figure}[t]
    \centering
    \includegraphics[width=0.6\linewidth]{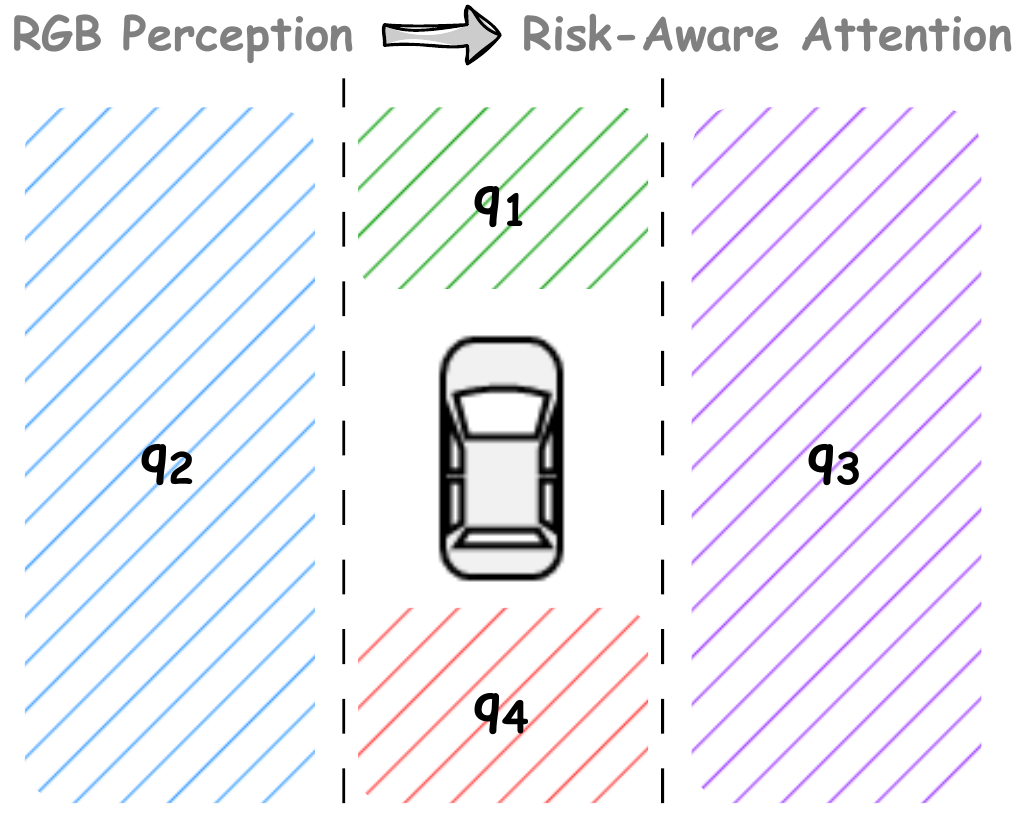}
    \caption{Spatial characterization of risk zones enhances the inference of VLM by enabling context-specific risk representations tailored to dynamic driving conditions.}
    \label{fig:risk_zones_demo}
\end{figure}
\begin{figure}[t]
    \centering
    \includegraphics[width=1\linewidth]{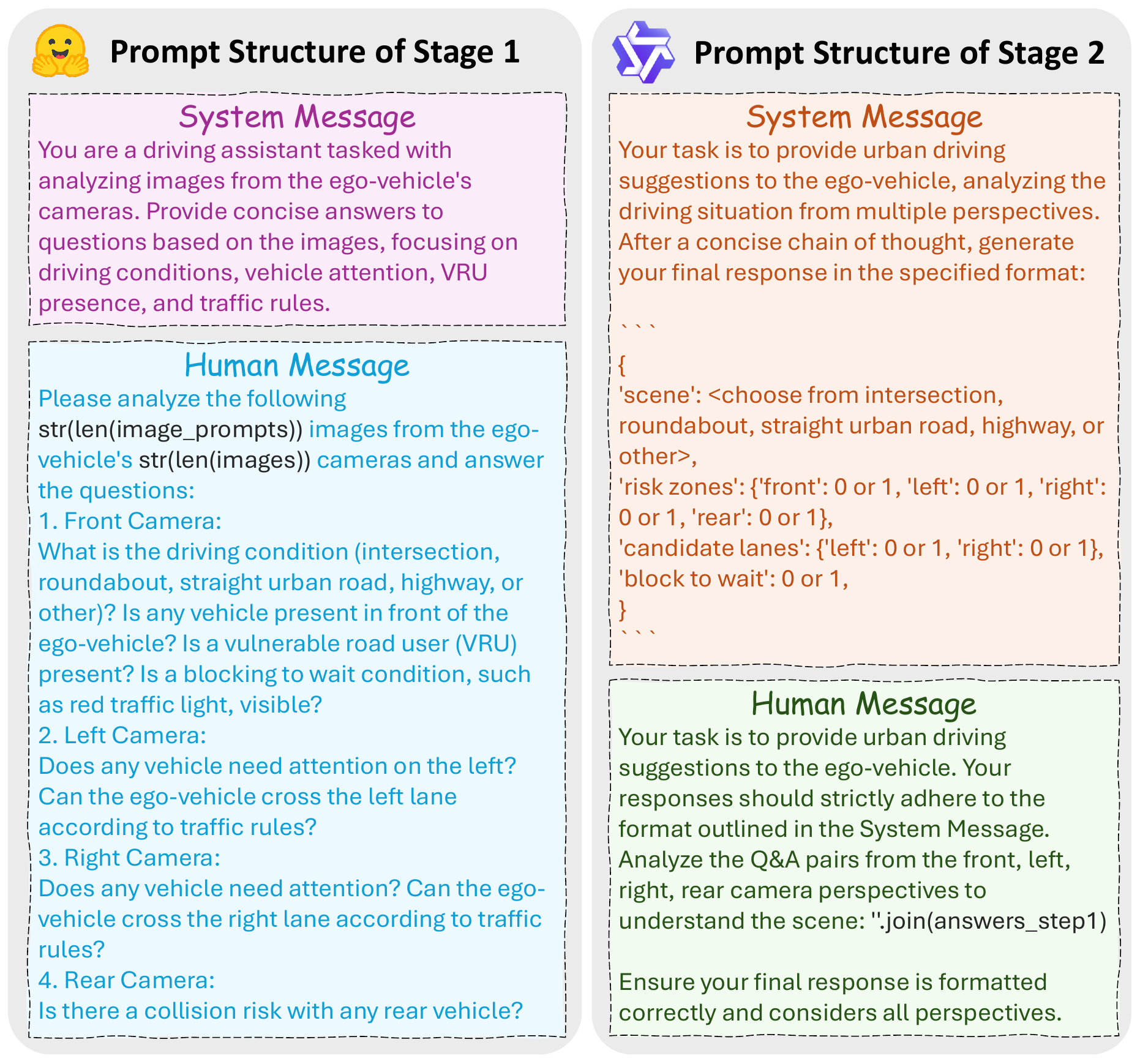}
    \caption{Structure of the prompts to the foundation models. The second stage prompt relies on the response of the first stage by retrieving and reformulating processes.}
    \label{fig:prompt_text_str}
\end{figure}
The system message $\mathcal{S}$ formalizes the foundation model's role and expected output through a structured template:
$
\mathcal{S} = (\mathcal{T}, \mathcal{F}),
$
where $\mathcal{T}$ is the task description, instructing the VLM to analyze camera images and the LLM to identify risk zones and critical objects (e.g., vehicles, VRUs, lane markings, traffic lights), $\mathcal{F}$ defines the response format as a dictionary $\mathcal{A}$ of attention objects:
\begin{equation}\label{eq:slow_response_stracture}
\begin{aligned}
\mathcal{A} = \big\{
&\text{scene}: \{s_1,, s_2, \ldots, s_p\},  \\
&\text{zones}: [q_1, q_2, \ldots, q_4], \\
&\text{mks}: [m_{\text{left}}, m_{\text{right}}], \\
&\text{btw}: \{0, 1\}
\big\},
\end{aligned}
\end{equation}
where $s_i$ is selected from {intersection}, {roundabout}, {straight urban road}, {highway}, and {others} for the scenario-specific reasoning of the second stage response. $q_i \in \mathbb{B}$ is a boolean indicator indicating whether a zone is at risk. The spatial delineation of risk zones is illustrated in Fig.~\ref{fig:risk_zones_demo}. 
$m_{\text{left/right}} \in \mathbb{B}$ indicates non-crossable and crossable lane markings of the left and right side as 0 and 1, respectively. In addition, $\text{btw}\in \mathbb B$ is a binary flag for blocking to wait, such as in the condition of encountering red traffic lights. 

\subsubsection{Human Message Representation}

The human message $\mathcal{H}^1$ for the VLM aggregates instant sensory data into a structured format:
\begin{equation}
\mathcal{H}^1 = (\mathcal{I}, \mathcal{T}^1),
\end{equation}
where $\mathcal{I}$ denotes the RGB images captured by onboard cameras from front, left, right, and rear directions, represented as tensors $\mathcal{I} \in \mathbb{R}^{n_c\times H \times W \times 3}$, in which $n_c$ is the number of the onboard cameras. As revealed in Fig.~\ref{fig:prompt_text_str}, $\mathcal{T}^1$ is the textual message providing specific commands and questions to the VLM regarding different perspectives of the autonomous vehicle.
Moreover, the second step of the foundation model inference requires the response of the VLM $\mathcal{A}^1$ to generate the human message $\mathcal{H}^2=\text{Rec}(\mathcal{H}^1,\mathcal{A}^1)$, where $\text{Rec}(\,\cdot\,)$ is the reconstruction procedure.
Along with the task description and formal response guidance in $\mathcal {S}^2$, the basic prompt for the VLM is ready to be invoked.

\subsubsection{Dialogue Pipeline with In-Context Learning}
To enhance contextual reasoning, VLM-UDMC retrieves dialogue examples $\mathcal{D} = \{(\mathcal{H}_i, \mathcal{A}_i)\}_{i=1}^{N_d}$ from memory storage, where $\mathcal{H}_i=[\mathcal{H}_i^1, \mathcal{H}_i^2]^\top$ is a historical human message and $\mathcal{A}_i=[\mathcal{A}_i^1, \mathcal{A}_i^2]^\top$ is the corresponding response from the foundation models. These examples are selected via similarity matching to the current scenario with their specific embeddings $e^v$ and $e^t$ generated by CLIP, a model trained on a vast amount of image-text pairs from the internet~\cite{radford2021learning}.
\begin{table}[t]
  \centering
  \caption{Parameters of the vehicle dynamics and potential functions of VLM-UDMC}
  \resizebox{1.0\linewidth}{!}{
    \begin{tabular}{cccccc}
    \toprule
    Notation & Value & Notation & Value & Notation & Value \\
    \midrule
    $k_f$ (\si{N/rad})       & -102129.83 & $k_r$ (\si{N/rad})       & -89999.98 & $l_f$ (\si{m})        & 1.287 \\
    $l_r$ (\si{m})           & 1.603      & $m$ (\si{kg})            & 1699.98   & $I_z$ (\si{kg\cdot m^2}) & 2699.98 \\
    $a_\text{NR}$              & 100.0      & $b_\text{NR}$              & 2.0       & $a_\text{CR}$            & 10.0 \\
    $b_\text{CR}$              & 0.5        & $a_\text{V}$               & 500.0     & $b_\text{V}$             & 1.0 \\
    $w_R$                      & 3.5        & $a_{\text{TL}_1}$             & 200.0     & $a_{\text{TL}_2}$           & 1000.0 \\
    $r_a$                      & 2.4        & $r_b$                      & 1.0       & $r_\text{offset}$       & 0.25 \\
    $\Phi_\text{front}$ (\si{deg})  & (0,0,0) & $\Phi_\text{left/right}$ (\si{deg})  & (25,0,90) & $\Phi_\text{rear}$ (\si{deg})  & (0,0,180) \\
    \bottomrule
    \end{tabular}%
  }
  \label{tab:params}
\end{table}

Therefore, incorporating the two steps of foundation models, the complete context generation process can be formalized as:
\begin{equation}
\mathcal{C}(t_m) = \left(
\mathcal{S},
\mathcal{H}(t_m),
\text{Retrieve}(\mathcal{M}, \mathcal{H}(t_m))
\right),
\end{equation}
where $\text{Retrieve}(\cdot)$ is a function that selects relevant historical examples from memory $\mathcal{M}$ based on similarity evaluation to the current human message $\mathcal{H}(t)$. The VLM processes this context to generate an updated attention dictionary $\mathcal{A}(t)$, which dynamically modifies the OCP's potential functions:
\begin{equation}
\begin{aligned}
F(\mathbf{P}(t), \mathbf{X}) = & \sum_{v \in \mathcal{V}(t)} F_{\text{V}}^v + \sum_{p \in \mathcal{P}(t)} F_{\text{VRU}}^p \\
& + \sum_{m \in \text{mks}} F_{\text{M}}^m + \text{btw} \cdot F_{\text{TL}},
\end{aligned}
\end{equation}
where $\mathcal{V}(t)$ and $\mathcal{P}(t)$ are the filtered vehicles and VRUs within the identified risk zones defined in (\ref{eq:slow_response_stracture}). Note that $F_M$ is equivalent to $F_\text{CR}$, as defined in~(\ref{eq:F_CR}), when $\text{mks} = 1$, and corresponds to $F_\text{NR}$, described in~(\ref{eq:F_NR}) otherwise. This structured approach ensures that the VLM's scene understanding is systematically translated into actionable control commands, enabling the autonomous vehicle to adapt to dynamic urban environments while maintaining computational efficiency.

\begin{table*}[htbp]
  \centering
  \caption{{Comparison of Driving Performance in Urban Driving Scenarios}}
    \begin{tabular}{cccccccc}
    \toprule
    \textbf{Method} & \textbf{Scenario} & \textbf{Comp. Time (ms)} & \textbf{Col.} & \textbf{TRV} & \textbf{IB} & \textbf{TTC Alarm Duration (s)} & \textbf{Travel Time (s)}\\
    \midrule
    \multirow{4}[2]{*}{Autopilot~\cite{dosovitskiy2017carla} (rule-based)} & ML-ACC & $10.61\pm3.57$ & 0     & 0     & 0     & 0.10 (0.4\%) & 26.6 \\
          & Roundabout & $8.93\pm2.85$ & 2     & 3     & 2     & 0.55 (1.5\%) & 35.6 \\
          & Intersection & $12.89\pm3.76$ & 1     & 0     & 1     & 0.65 (3.6\%) & 18 \\
          & Mix-T-Junction & $5.14\pm0.64$ & 1     & 0     & 0     & \textbackslash{} & \textbackslash{} \\
    \midrule
    \multirow{4}[2]{*}{InterFuser~\cite{shao2023safety} (learning-based)} & ML-ACC &  $178.20\pm 23.78$    &  0     &  0     & 0     &  0 (0.0\%)       &   52.9 \\  
          & Roundabout &  $179.98\pm 22.62$    &  0     &   0    &    0   & 0 (0.0\%)      &  52.85 \\ 
          & Intersection &     $180.33\pm 25.28$     &   0    &   0    &   0       & 0 (0.0\%)       &  28.95\\
          & Mix-T-Junction &  $162.74\pm 30.41$     &    0   &   0    &   0   &   0 (0\%)      & 45.75 \\
    \midrule
    \multirow{4}[2]{*}{RSS~\cite{shalev2017formal} (reactive-based)} & ML-ACC &   $11.62\pm3.14$    & 0     & 0     & 0     & 0.15 (0.6\%) & 26.7 \\
          & Roundabout &  $9.67\pm3.22$  & 1     & 0     & 0     & 0.85 (2.3\%) & 37.75 \\
          & Intersection &   $13.29\pm3.48$    & 0     & 0     & 0     & 0.20 (1.0\%) & 20.5 \\
          & Mix-T-Junction &   $6.49\pm 9.81$    & 1     & 2     & 0     & 0 (0.0\%) & 31.25 \\
    \midrule
    \multirow{4}[2]{*}{UDMC~\cite{liu2025udmc} (optimization-based)} & ML-ACC & $35.07\pm13.00$ & 0     & 0     & 0     & 0 (0.0\%) & 16.65 \\
    & Roundabout & $30.35\pm12.68$ & 0     & 0     & 0     & 0 (0.0\%) & 17.8 \\
    & Intersection & $25.44\pm14.32$ & 0     & 0     & 0     & 0 (0.0\%) & 16.5 \\
    & Mix-T-Junction & $27.32\pm11.34$ & 0     & 0     & 0     & 0 (0.0\%) & 24.75 \\
    \midrule
    \multirow{4}[2]{*}{VLM-UDMC (Ours)} & ML-ACC & $9.10\pm4.28$ & 0     & 0     & 0     & 0 (0.0\%) & 15.5 \\
    & Roundabout & $12.74\pm5.79$ & 0     & 0     & 0     & 0 (0.0\%) & 17.75 \\
    & Intersection & $9.81\pm6.67$ & 0     & 0     & 0     & 0.05 (0.3\%) & 15.1 \\
    & Mix-T-Junction & $12.11\pm6.16$ & 0     & 0     & 0     & 0 (0.0\%) & 22.60 \\
    \bottomrule
    \end{tabular}%
    \begin{flushleft}
        {\small \textbf{Note:} ML-ACC denotes the multilane adaptive cruise control scenario, while Mix-T-Junction represents the T-junction driving scenario with mixed traffic conditions. Besides, Col, TRV, and IB represent the counts of collisions, traffic rule violations, and impolite behaviors, respectively. TTC Alarm Duration (s) indicates the duration that the Time-To-Collision (TTC) value remains below 1.5\,s.}
    \end{flushleft}
  \label{tab:CompScenariosMethods}%
\end{table*}%
\section{Experimental Results and Analysis}
In this section, we evaluate VLM-UDMC in both simulated and real-world driving scenarios to assess its scene understanding and risk mitigation capabilities. We compare our framework with representative baselines across diverse driving scenarios, evaluating driving safety and efficiency through both quantitative and qualitative measures. We also perform an ablation study on key features of VLM-UDMC before assessing its reasoning performance in real-world driving scenarios.

\subsection{Environmental Settings}
All experiments are conducted on a high-performance workstation equipped with an AMD Ryzen Threadripper Pro 7975WX processor, featuring 32 cores and 64 threads at 3.5 G\,Hz. The system is configured with 512 GB of DDR5 RAM to support large-scale data processing and multitasking. For accelerated computation, the workstation includes two NVIDIA RTX 6000 Ada Generation GPUs, each providing 48 GB of GDDR6 memory. 
The software environment was based on Ubuntu 22.04 LTS operating system, with CUDA 12.1 and cuDNN 8.9 for GPU acceleration. All models were implemented using Python 3.9, with deep learning frameworks including PyTorch 2.4 and training scripts optimized for multi-GPU execution.
The dynamically formulated OCP is solved using the Interior Point Optimizer (IPOPT) with CasADi~\cite{Andersson2019}.
In the CARLA 0.9.15 simulator~\cite{dosovitskiy2017carla}, the autonomous vehicle is controlled by throttle, brake, and steering, which are transformed by a PID controller with a low-pass filter for a smoother control input from the acceleration and steering values generated by MPC. The surrounding vehicles are driven by the embedded \texttt{autopilot} for a closed-loop evaluation. The relevant parameters of the VLM-UDMC utilized in the experiments are exhibited in Table~\ref{tab:params}, which is of general utility in urban driving scenarios. $\Phi_j$ denotes the spatial configuration with the order of (pitch, roll, yaw) of the $j$-th onboard camera.

\subsection{Comparison and Effective Analysis}\label{subsec:comparison}
To evaluate the rationality of scene understanding and risk attention generation for available potential functions, we compare our proposed urban driving algorithm with a variety of baselines, including rule-based, optimization-based, and learning-based methods, including Autopilot~\cite{dosovitskiy2017carla}, RSS~\cite{shalev2017formal}, InterFuser~\cite{shao2023safety}, and UDMC~\cite{liu2025udmc}. All methods are deployed in identical driving scenarios, including multilane adaptive cruise control, roundabout navigation, intersection crossing, and T-junction driving with mixed traffic conditions. During evaluation, surrounding traffic participants are consistently spawned at the same points across different trials, and they react to the behavior of the ego vehicle to simulate realistic interactions.
\begin{table}[t]
\centering
\caption{Comparison of Model Performance and Total Inference Time}
\label{tab:model_comparison}
\begin{tabular}{c c c c c}
\toprule
\textbf{Model} & \textbf{MSE} & \textbf{RMSE} & \textbf{MAE} & \textbf{Infer. Time (ms)} \\
\midrule
SGPR~\cite{liu2024incremental} & 9.82 & 3.13 &\underline{0.39} & 6.29 \\
LSTM & 1.25& 1.12& 0.43& \underline{0.18}\\
Dlinear & 1.41& 1.19& \textbf{0.34}& \textbf{0.17}\\
Transformer & \underline{1.10}&
\underline{1.05}&
0.46&
0.43\\
Ours & \textbf{0.80}& \textbf{0.90}& \textbf{0.34} & 0.27\\
\bottomrule
\end{tabular}
\begin{flushleft}
    {\small \textbf{Note:} Best results are in \textbf{bold}, second best ones are with \underline{underline}.}
\end{flushleft}
\label{tab:predict_comp}
\end{table}
\begin{figure*}[htbp]
    \centering
    \subfigure[Multiple lane adaptive cruise control with a stop sign]{
        \includegraphics[width=0.45\linewidth]{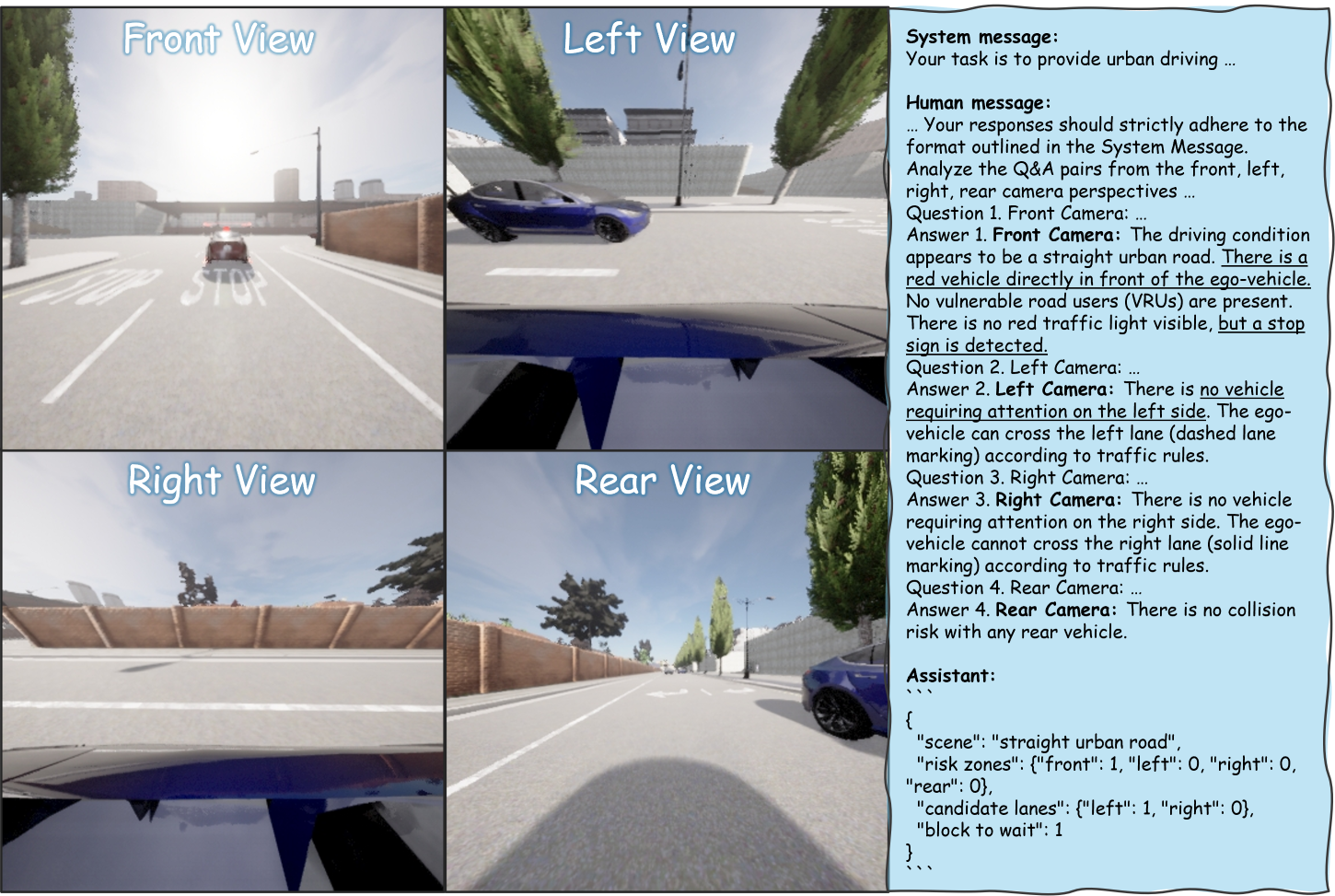}
        \label{fig:mlACC_dialogue}
    }
    \subfigure[Roundabout with a front vehicle changing lane]{
        \includegraphics[width=0.45\linewidth]{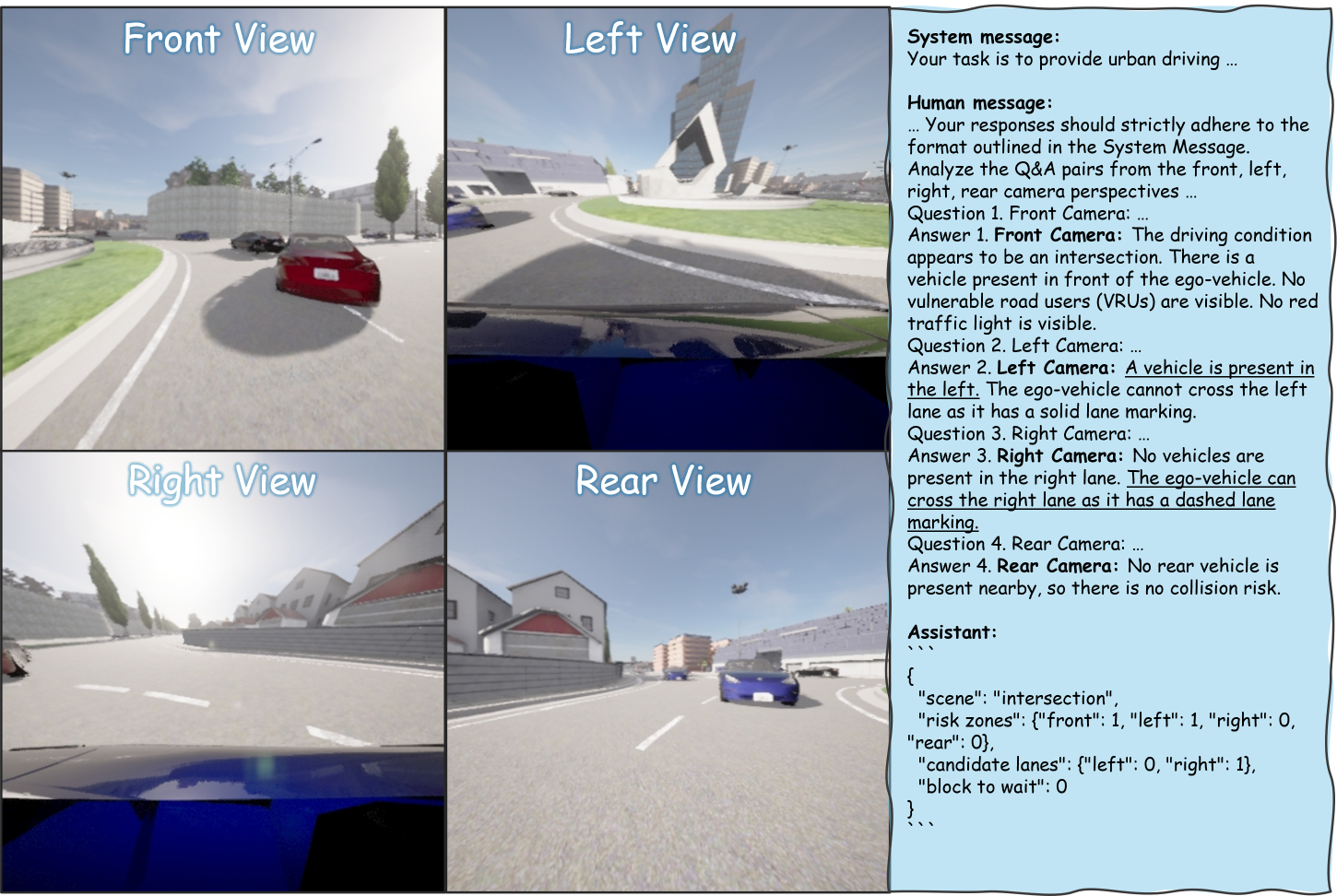}
        \label{fig:roundabout_dialogue}
    }
    \subfigure[Crossroad with the ego vehicle is at the center of the intersection]{
        \includegraphics[width=0.45\linewidth]{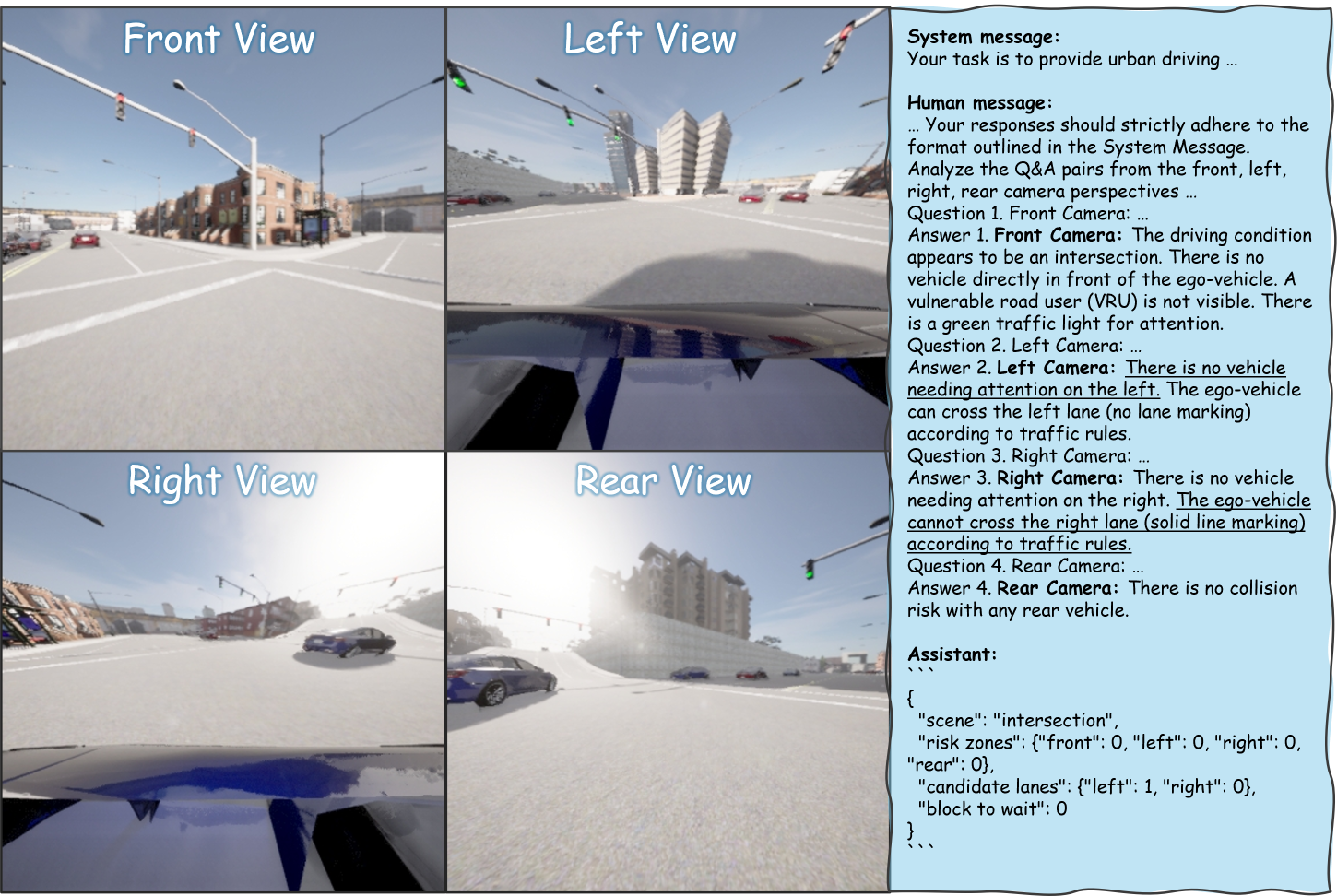}
        \label{fig:crossroad_dialogue}
    }
    \subfigure[T-junctions involving VRUs]{
        \includegraphics[width=0.45\linewidth]{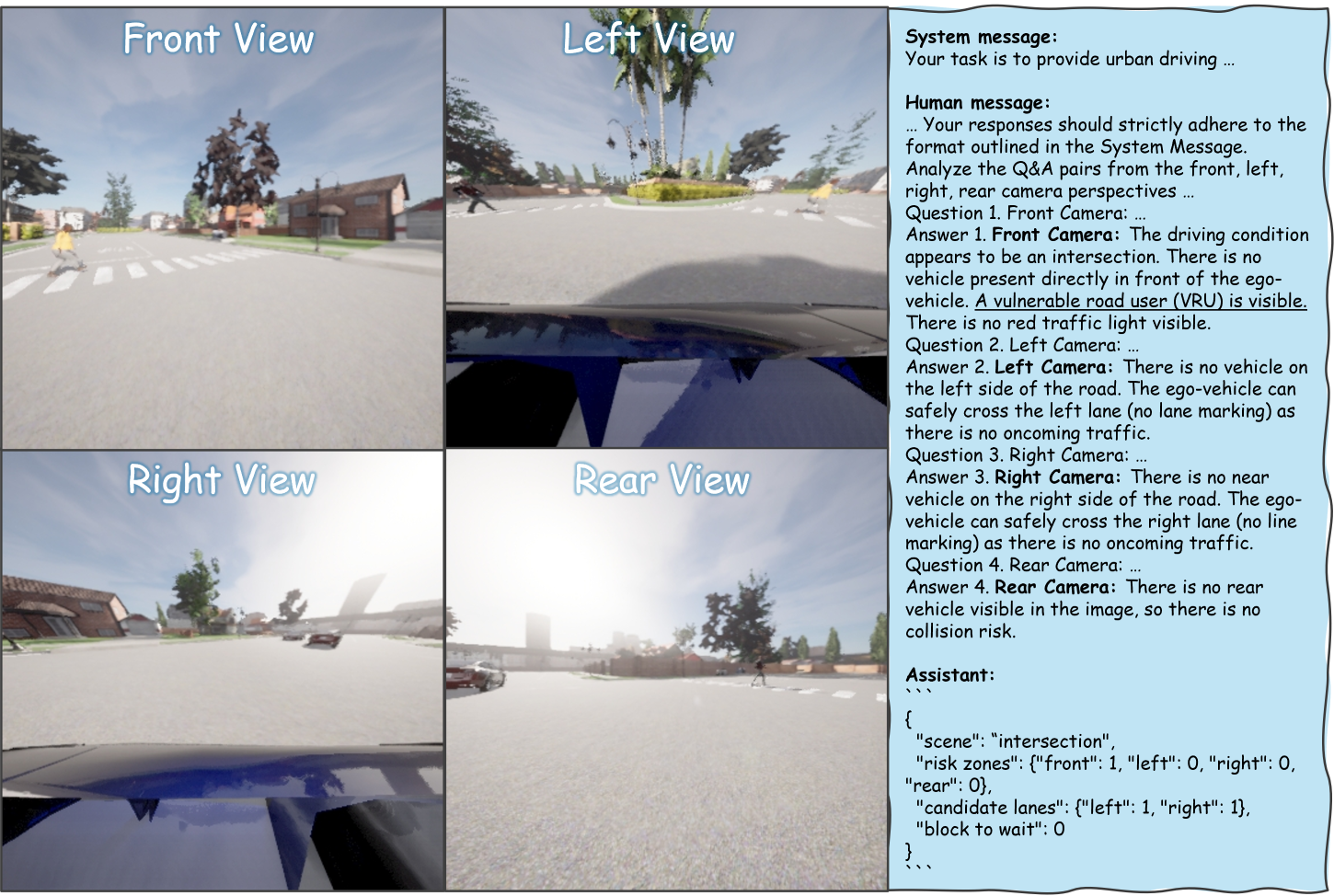}
        \label{fig:mixedtraffic_dialogue}
    }
    \caption{Demonstration of the multimodal dialogues during urban driving. The vision data contains the images from different perspectives, and the key insights of the textual dialogue are \underline{underlined} for clear illustration.}
    \label{fig:all_scenarios_dialogue}
\end{figure*}
\begin{figure}[t]
\centering
\subfigure[Multiple lane adaptive cruise control scenario]{
\includegraphics[width=1\linewidth]{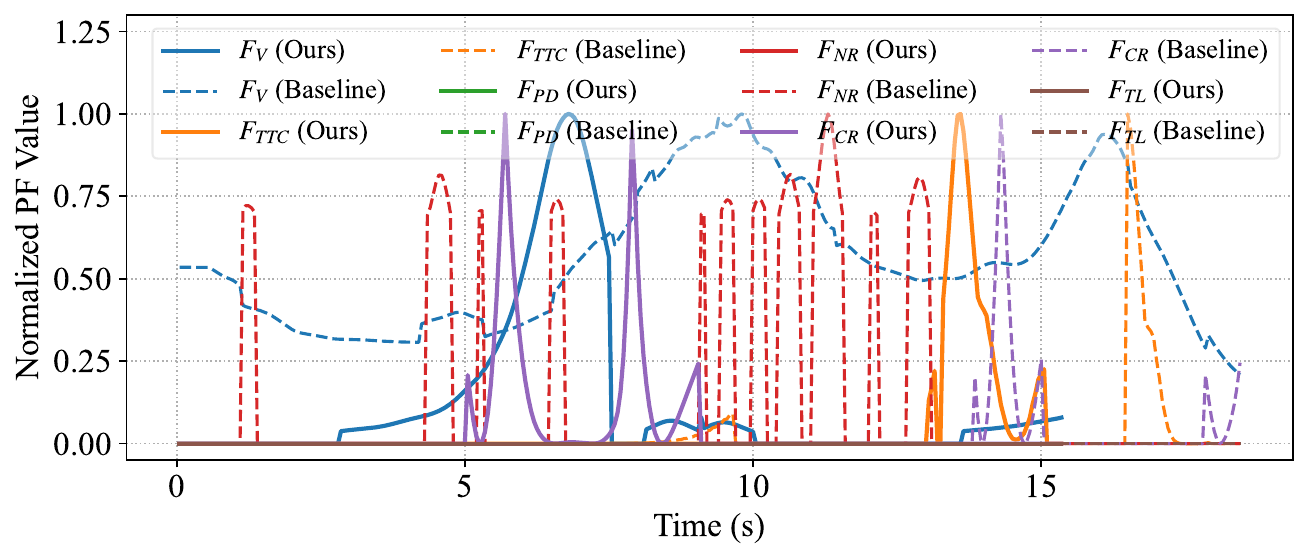}
\label{fig:mlACC_PF}
}
\subfigure[Roundabout scenario]{
\includegraphics[width=1\linewidth]{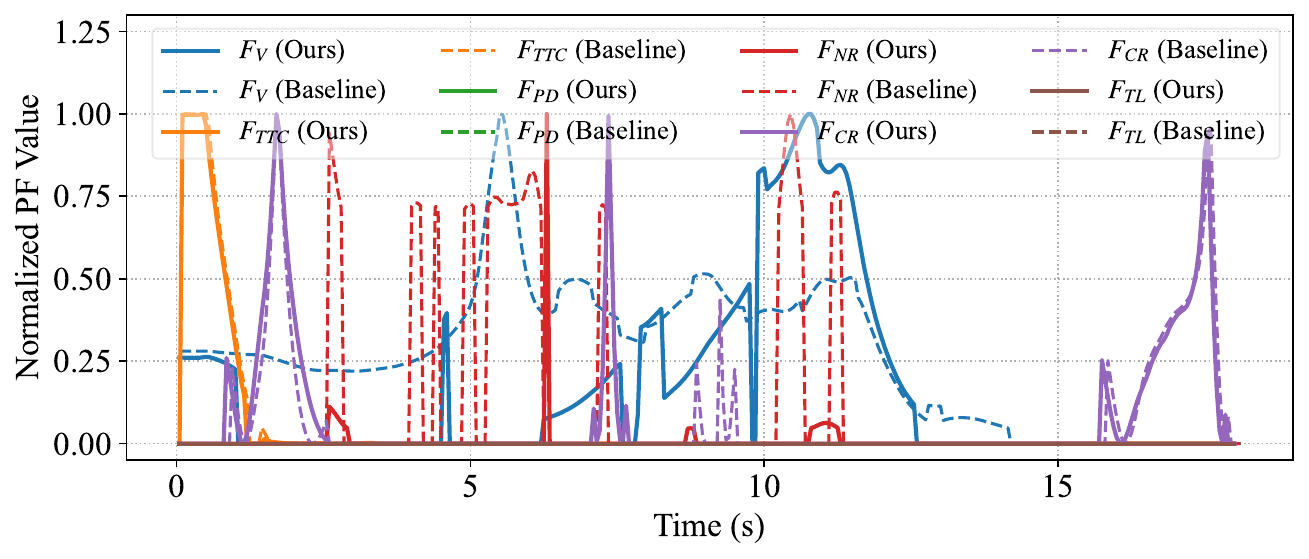}
\label{fig:roundabout_PF}
}
\subfigure[Crossroad scenario]{
\includegraphics[width=1\linewidth]{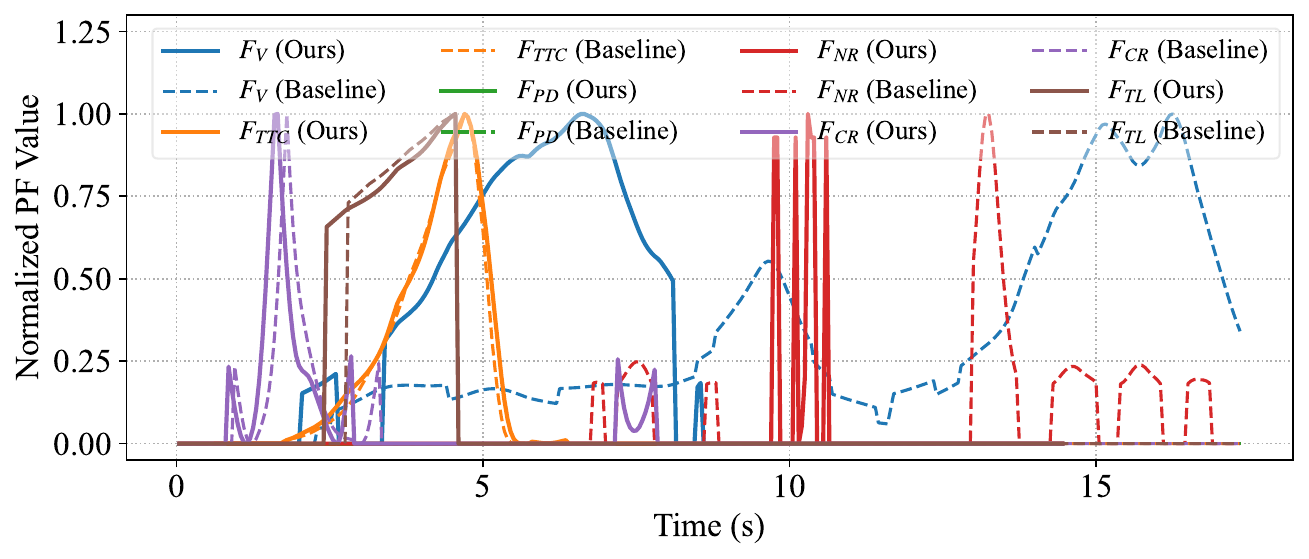}
\label{fig:crossroad_PF}
}
\subfigure[Mixed traffic scenario]{
\includegraphics[width=1\linewidth]{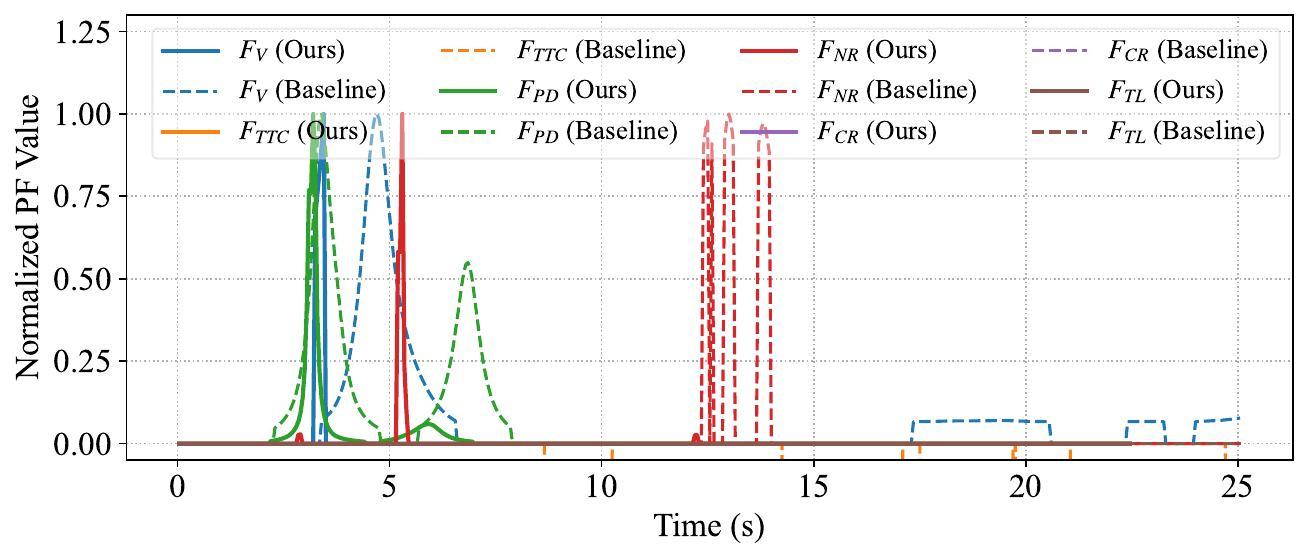}
\label{fig:mixedtraffic_PF}
}
\caption{The value of the assigned potential functions during different driving scenarios compared with the baseline values.}
\label{fig:all_scenarios_PF}
\end{figure}
\begin{figure}[tbp]
    \centering
    \subfigure[Multiple lane adaptive cruise control scenario]{
        \includegraphics[width=1.0\linewidth]{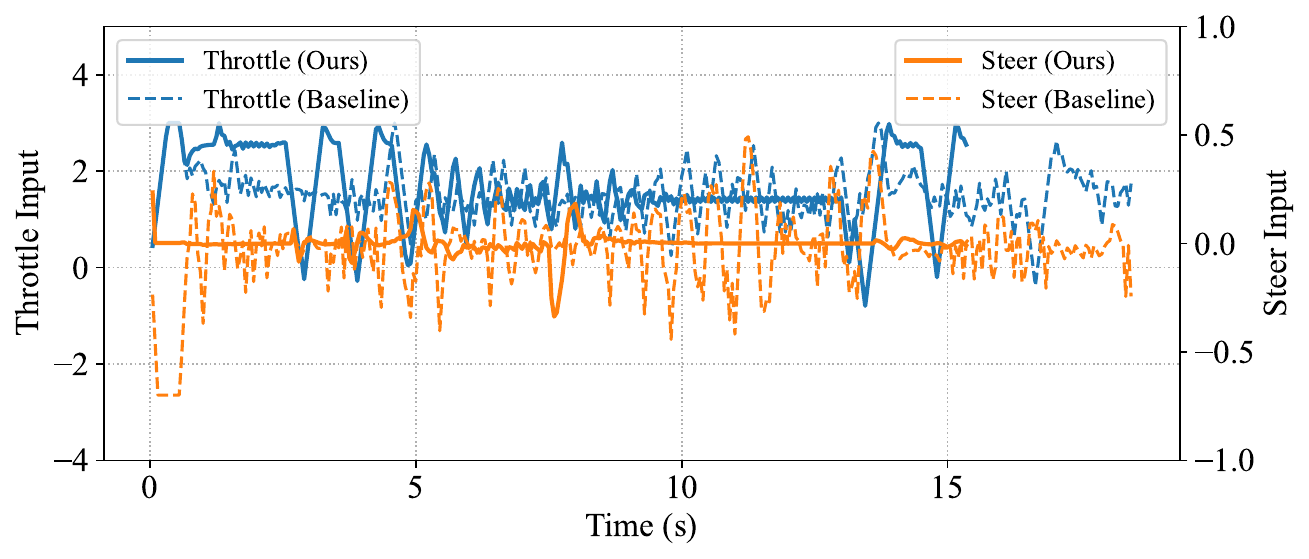}
        \label{fig:mlACC_control_comp}
    }
    \subfigure[Roundabout scenario]{
        \includegraphics[width=1.0\linewidth]{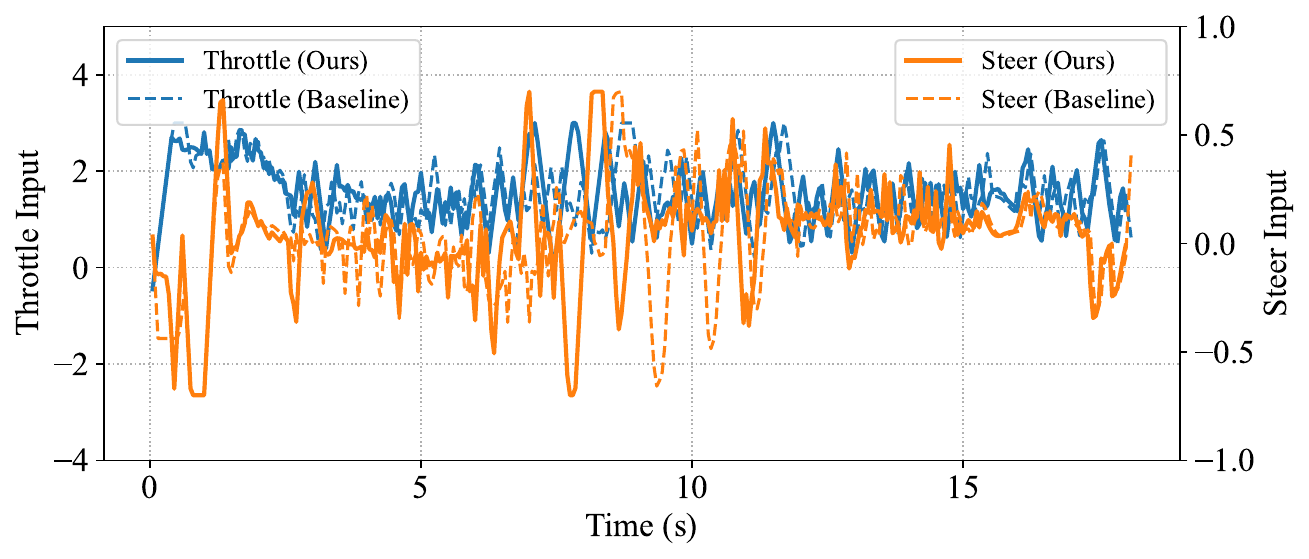}
        \label{fig:roundabout_control_comp}
    }
    \subfigure[Crossroad scenario]{
        \includegraphics[width=1.0\linewidth]{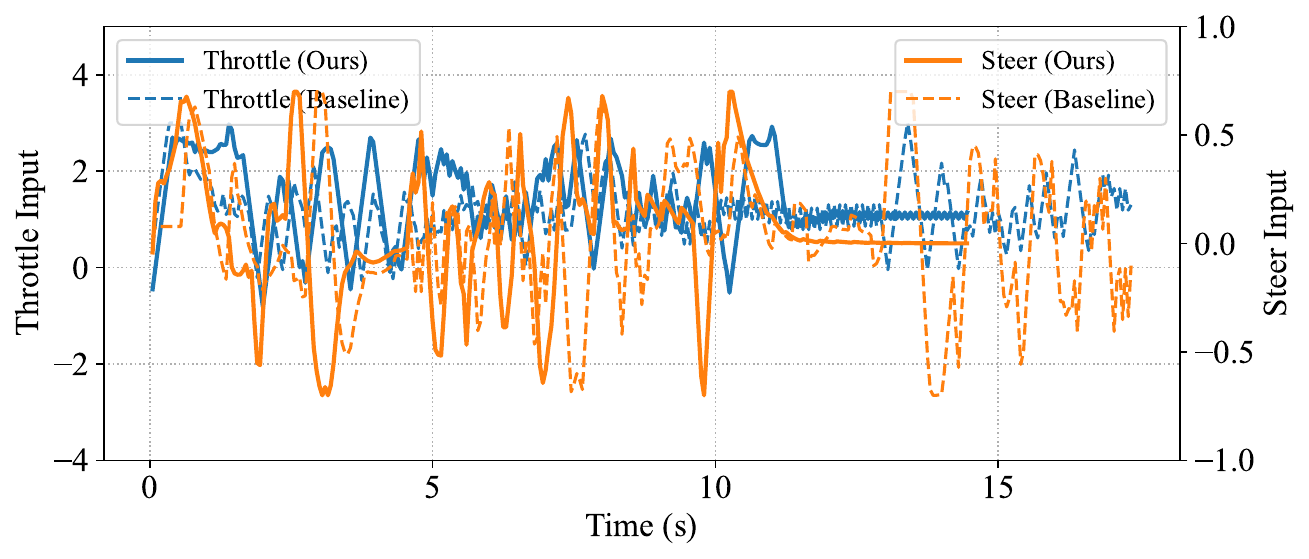}
        \label{fig:crossroad_control_comp}
    }
    \subfigure[Mixed traffic scenario]{
        \includegraphics[width=1.0\linewidth]{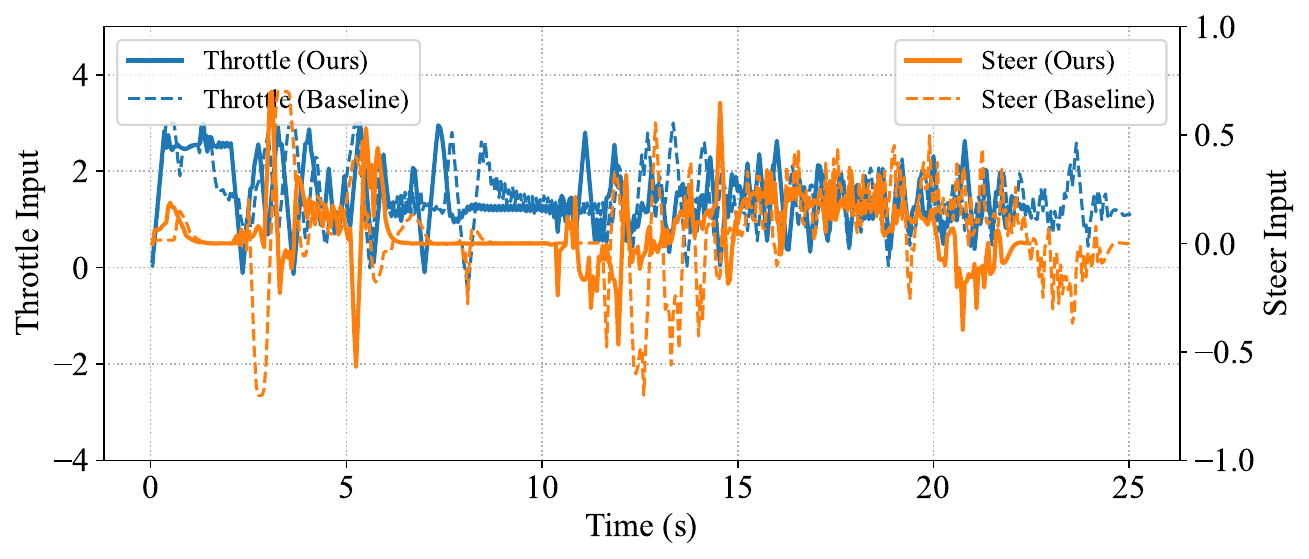}
        \label{fig:mixedtraffic_control_comp}
    }
    \caption{Comparison of control inputs across different driving scenarios between the proposed method and the baseline.}
    \label{fig:all_scenarios_control_inputs}
\end{figure}
The simulation results in Table~\ref{tab:CompScenariosMethods} demonstrate that our proposed VLM-UDMC achieves the highest traffic efficiency across all four representative driving scenarios, while maintaining robust safety in interactions with other traffic participants. The fast-slow structure of VLM-UDMC allows for the neglect of unnecessary traffic participants in the adaptive OCP, thereby enhancing computational efficiency. However, it is important to note that a time-to-collision alarm briefly occurred during the intersection scenario due to the oversight of a vehicle crossing at the same time.

In terms of the performance of the proposed traffic participant prediction module, as shown in Table~\ref{tab:predict_comp}, the proposed multi-kernel prediction approach consistently achieves superior performance across all evaluation metrics compared to existing baselines. Specifically, our method attains the lowest MSE, RMSE, and MAE, representing improvements of 27.3\% in MSE and 14.3\% in RMSE over the second-best Transformer model. For MAE, our method reduces the error by 25.6\% compared to the Transformer.
In addition to its state-of-the-art accuracy, our method maintains a highly competitive inference speed of 0.27 ms per batch, which is 37.2\% faster than the Transformer (0.43 ms). Reaching inference speeds on the millisecond scale (under 0.3 ms) guarantees that the model will not hinder downstream planning and control processes. Consequently, the inference efficiency of our model is on par with that of DLinear and LSTM. Moreover, compared to the traditional SGPR method, our method is not only substantially more accurate but also over 23 times faster in inference. 
The results indicate that our proposed prediction approach provides an optimal balance between accuracy and computational efficiency, making it particularly suitable for real-time or resource-constrained trajectory prediction tasks. This capability is crucial for enabling the low-level OCP to deliver vehicle control commands in real-time.

Qualitative analysis of the decisions made by VLM-UDMC's slow system is conducted to assess its effectiveness in scene understanding and rational decision-making. Fig.~\ref{fig:all_scenarios_dialogue} presents representative dialogue segments generated during adaptive cruise control, roundabout navigation, intersection crossing, and T-junction driving scenarios. In Fig.~\ref{fig:mlACC_dialogue}, as the ego vehicle approaches the end of a straight urban road, VLM-UDMC detects a blocking vehicle and a stop sign ahead. After conducting relevant analysis, the slow system commands the absorption of traffic participants from the front zone, suggesting a lane change to the left. Although a vehicle is present in the left lane, VLM-UDMC disregards it, as there are no obstacles ahead and the current lane is deemed safe for the ego vehicle. Conversely, in the roundabout scenario depicted in Fig.~\ref{fig:roundabout_dialogue}, VLM-UDMC's fast system responds by incorporating both front and left risk zones into the OCP for attention. Additionally, when the ego vehicle turns right at an intersection, as shown in Fig.~\ref{fig:crossroad_dialogue}, VLM-UDMC detects no nearby traffic participants moving toward the ego vehicle, identifying no risk zones and permitting a left-turn maneuver. Finally, in Fig.~\ref{fig:mixedtraffic_dialogue}, several VRUs are present on pedestrian crossings, and VLM-UDMC focuses attention on the front zone within the OCP, as other pedestrian routes do not intersect with the ego vehicle's path.
Regarding the effectiveness of the fast system of VLM-UDMC, the rational selection of risk zones for attention, as depicted in Fig.~\ref{fig:all_scenarios_PF}, reduces the sum of potential functions, thereby decreasing the complexity of the OCP. This feature facilitates smoother and more consistent spatial-temporal motion of the ego vehicle. As illustrated in Fig.~\ref{fig:all_scenarios_control_inputs}, across four representative urban scenarios, the control inputs, including acceleration and steering angles, of our proposed VLM-UDMC are notably smoother, with fewer action adjustments compared to the baseline. UDMC incorporates all surrounding traffic participants into the OCP, leading to a more complex optimization problem with more local optima.
\subsection{Ablation Study}
In this part, we conduct an ablation study to evaluate the key features of the proposed VLM-UDMC framework, focusing on its application in complex urban scenarios, specifically intersection crossings. The features under examination include the two-step reasoning mechanism, risk zone representation, and memory-based in-context learning.

As illustrated in Table~\ref{tab:ablation}, the VLM-UDMC variant lacking all three features exhibits twice the number of collisions with surrounding vehicles. This is attributed to the inadequate scenario understanding provided by the one-step reasoning process, which fails to accurately interpret the directions of relevant traffic objects. Additionally, this version records high counts of traffic rule violations and impolite behaviors, culminating in unsafe driving performance.
Introducing the two-step reasoning mechanism enhances scene comprehension, thereby improving driving safety and efficiency. However, the inherent latency in response from the foundational models results in delayed updates of vehicle IDs, which are crucial for the adaptive OCP.
The integration of risk zone representation into the ablated version significantly enhances urban driving performance, eliminating collisions and impolite behaviors entirely. Finally, the complete version of VLM-UDMC, which incorporates all features, demonstrates no traffic rule violation and achieves the shortest travel time during intersection crossings. This underscores the effectiveness of the memory-based in-context learning strategy in optimizing driving performance.
\begin{table}[t]
  \centering
  \caption{Ablation Results of Key Features of VLM-UDMC}
    \begin{tabular}{cccccccc}
    \toprule
    \multicolumn{3}{c}{Ablation Items} & \multirow{2}{*}{Col.} & \multirow{2}{*}{TRV} & \multirow{2}{*}{IB} & \multirow{2}{*}{TTC Alarm} & \multirow{2}{*}{Travel Time} \\
    \cmidrule{1-3}
     TS & RZ & MEM & & & & & \\ 
    \midrule
    \ding{55} & \ding{55} & \ding{55} & 2   & 4   & 3   & 0.1\,s   & 43.15\,s \\
    \ding{51} & \ding{55} & \ding{55} & 1   & 2   & 2   & 0.8\,s   & 16.6\,s  \\
    \ding{51} & \ding{51} & \ding{55} & 0   & 1   & 0   & 0.05\,s  & 15.3\,s  \\
    \ding{51} & \ding{51} & \ding{51} & 0   & 0   & 0   & 0.05\,s  & 15.1\,s  \\
    \bottomrule
    \end{tabular}%
  \begin{flushleft}
    {\small \textbf{Note:} TS, RZ, and MEM denote Two-Step Reasoning, Risk Zone, and Memory-based In-Context Learning, respectively.}
  \end{flushleft}
  \label{tab:ablation}%
\end{table}%
\subsection{Field Experiment with a Physical Autonomous Vehicle}
\begin{figure}[t]
\centering
\includegraphics[width=1\linewidth]{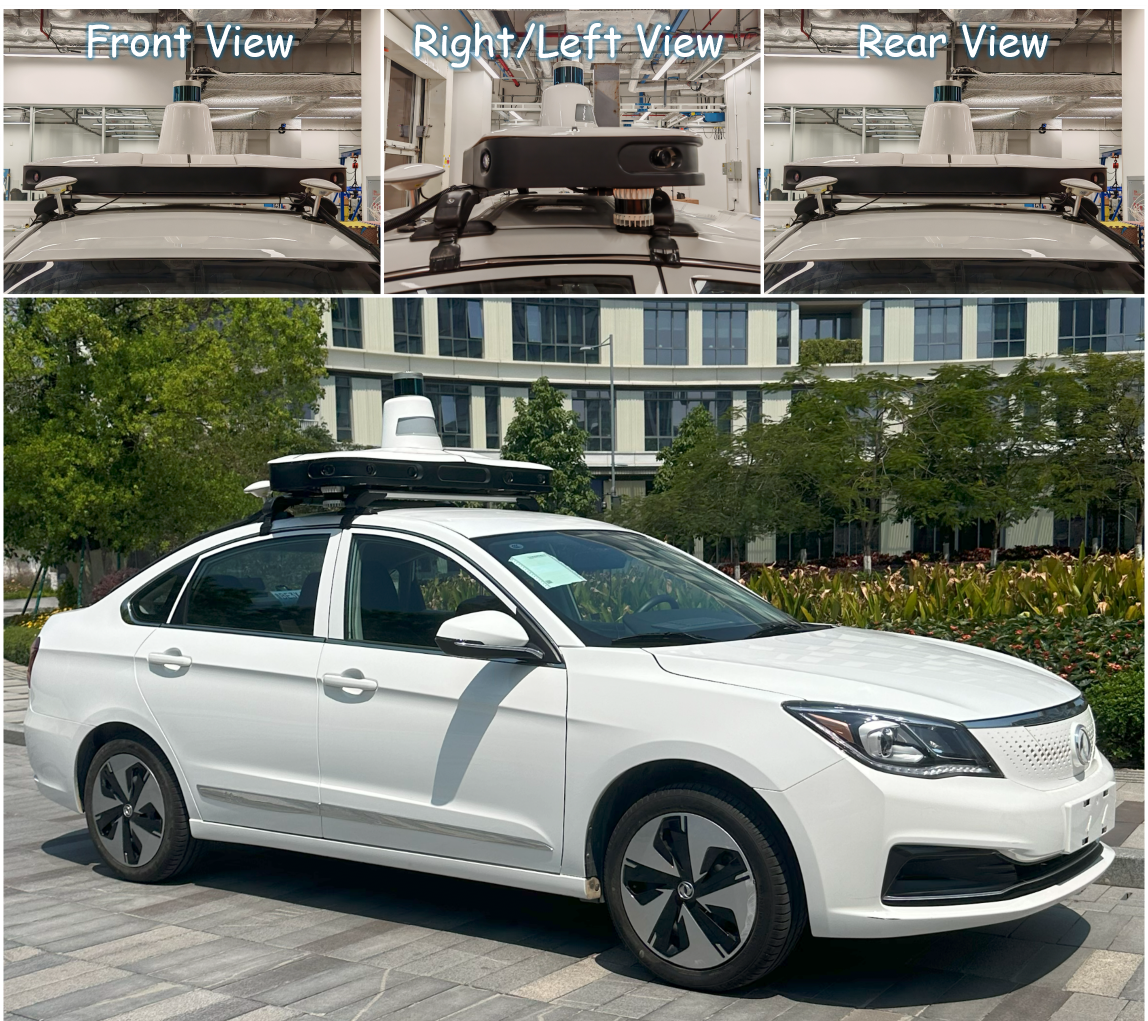}
\caption{The perception system of the intelligent vehicle. It contains six RGB cameras mounted on the top of the vehicle to capture the complete surround view for scene understanding and risk attention generation.}
\label{fig:hardware_vehicle}
\end{figure}
The scene understanding capabilities of our proposed urban driving framework are evaluated using real-world driving visual data. For this assessment, we conduct the autonomous driving process on a university campus. Instantaneous RGB images are downsampled and processed by locally deployed SmolVLM and QWen3-4B foundation models. These models use the two-step reasoning strategy to provide responses $\mathcal{A}$ regarding risk zones and other potential functions related to traffic rules from the slow system of VLM-UDMC.

As illustrated in Fig.~\ref{fig:hardware_vehicle}, the perception system of the intelligent vehicle is engineered through the integration of six RGB cameras, which are strategically positioned on the vehicle's roof. This arrangement facilitates the acquisition of a comprehensive surround view. The RGB cameras capture images at a frequency of 10\,Hz, which allows for near real-time perception.  It is important to note that the camera configuration of the real vehicle differs slightly from our proposed simulated camera setup. Consequently, we made minor modifications to the prompt in the first stage of VLM reasoning to accommodate the new camera configuration.

\begin{figure}[t]
    \centering
    \subfigure[Straight urban road driving with an oncoming vehicle]{
        \includegraphics[width=0.9\linewidth]{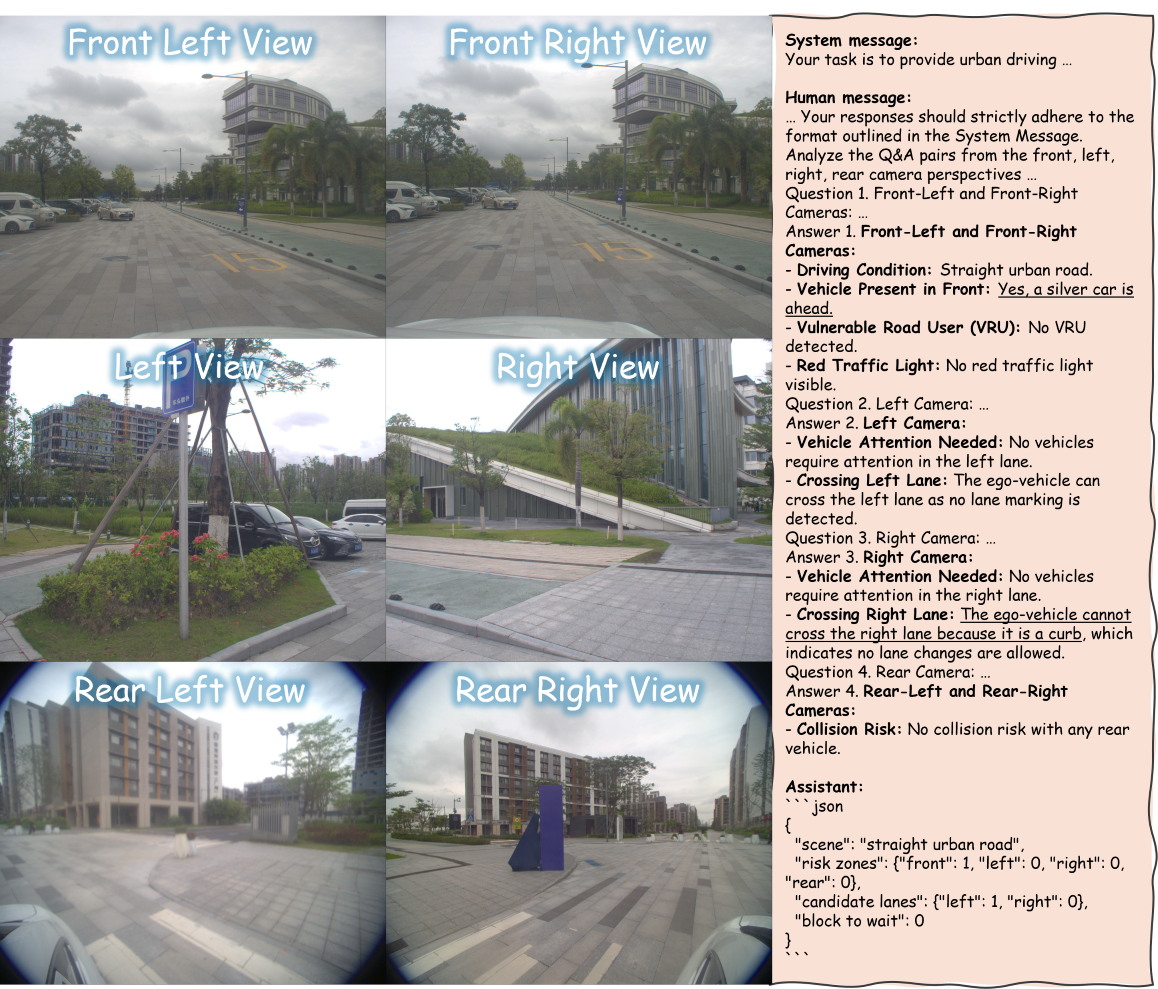}
        \label{fig:real_oncoming_veh_dialogue}
    }
    \subfigure[Straight urban road driving with a VRU on the side]{
        \includegraphics[width=0.9\linewidth]{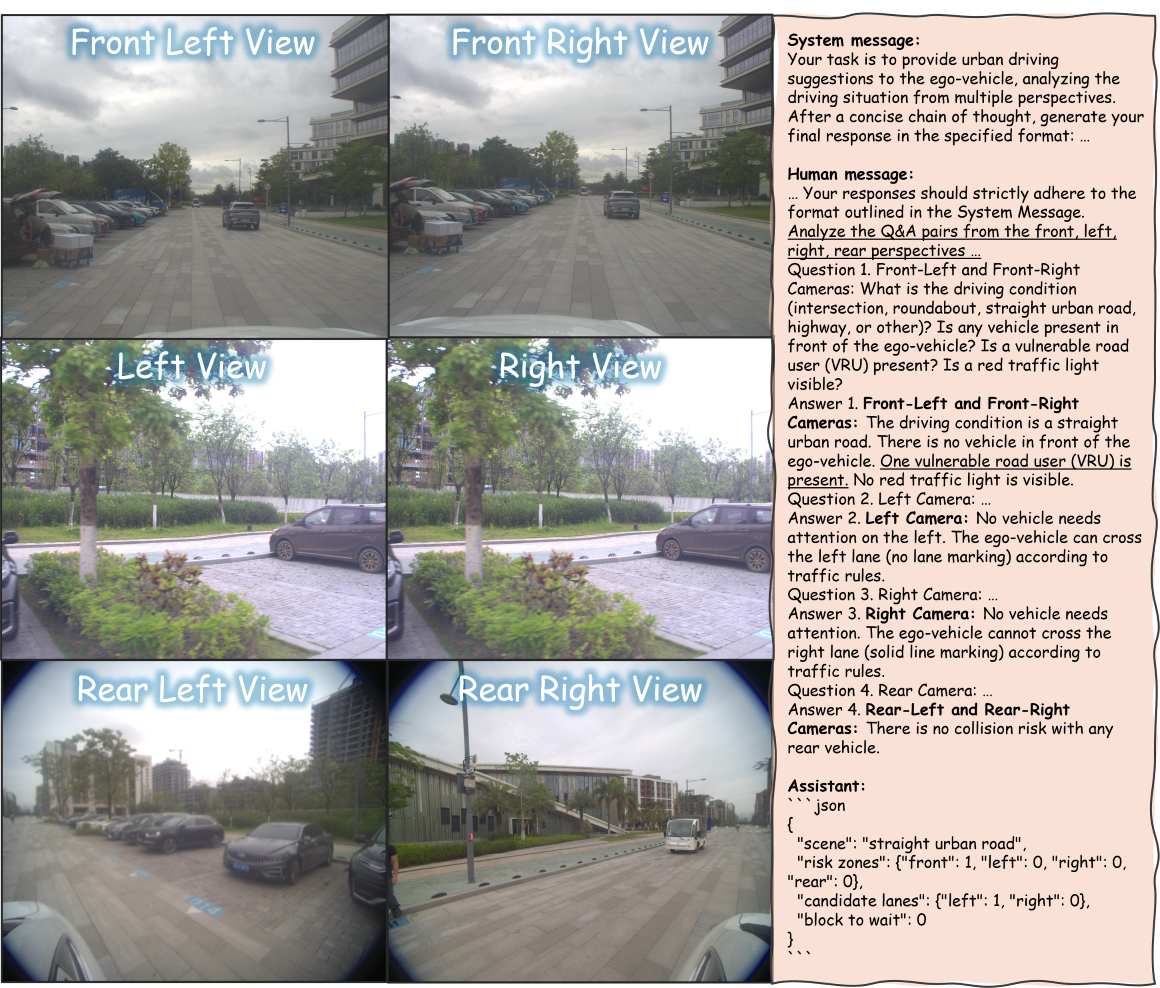}
        \label{fig:real_vru_dialogue}
    }
    \caption{Multimodal dialogue demonstration during campus driving, with vision data from six cameras. Key textual insights are \underline{underlined} for clarity.}
    \label{fig:real_world_dialogue}
\end{figure}

We present several response instances from VLM-UDMC to demonstrate the effectiveness of our proposed two-stage scene understanding mechanism and risk zone description strategy. In Fig.~\ref{fig:real_oncoming_veh_dialogue}, a silver sedan exits a parking lot and approaches the ego vehicle. VLM-UDMC accurately identifies this situation, marking the front zone as at risk and incorporating all detected obstacles in front of the ego vehicle into the fast system. Additionally, VLM-UDMC recognizes the road curb on the right and advises against crossing into the right lane. Conversely, in another driving condition depicted in Fig.~\ref{fig:real_vru_dialogue}, the front vehicle is not deemed an object of attention. Instead, a person working near a car trunk is identified as a VRU, activating the corresponding risk zone in the final response $\mathcal{A}$. The above analysis indicates that our proposed VLM-UDMC can adapt to different sensor configurations from simulation to real-world situations, while maintaining effective scene understanding and risk zone reasoning capabilities.

\section{Discussion and Conclusion}
\subsection{Discussion}
The proposed urban driving framework leverages foundation models with an adaptive OCP to achieve human-like scene understanding and the subsequent decision-making and motion control. The foundation models, enhanced by two-step reasoning and RAG for in-context learning, leverage relevant driving dialogues to generate risk zone-based insights that directly inform the low-level MPC. Unlike previous parameter-based adjustments by VLMs, this semantic guidance enables dynamic adaptation of control objectives, ensuring safe and compliant behaviors in complex scenarios such as left-turning while yielding to oncoming traffic.
The presented multi-kernel prediction mechanism addresses the need for real-time motion prediction by decomposing trajectories into trend and remainder components with outstanding predictive efficiency. This integration with the fast system of VLM-UDMC improves the rationality of decision-making and motion control, enabling the autonomous vehicle to better anticipate and respond to the movements of surrounding agents.

However, we observe that the performance of VLM-UDMC is highly influenced by prompt engineering. For instance, sending images in their original size to the VLM results in significantly longer inference times and may lead to inaccurate responses. Additionally, the system message must be carefully designed to ensure stable and pertinent responses.
Furthermore, due to the inherent limitations of foundation models, we are restricted to using only two modalities, vision and language, for scene understanding. This constraint limits the VLM's ability to gain accurate spatial insights. Lastly, the training-free and rapid retrieval capabilities of RAG present an exciting opportunity for lifelong learning in autonomous driving, making it a promising area for future research.
\subsection{Conclusion}
In this work, we propose a VLM-enhanced UDMC framework with multiple time-scales for urban autonomous driving that replicates human-like attention capabilities while maintaining transparency and interpretability.
Within this unified framework, the upper decision-making system seamlessly reconfigures the OCP by adaptively invoking the potential functions representing the traffic factors.
To enhance scene understanding, the proposed two-step reasoning policy with RAG utilizes foundation models to process multimodal inputs and retrieve contextual knowledge, enabling the generation of grounded risk-aware insights. This mechanism also promotes greater inference consistency, particularly in long-tail situations.
Meanwhile, the motion control system employs a lightweight multi-kernel decomposed LSTM to proactively anticipate future trajectories of surrounding agents by extracting smooth trend representations within the planning horizon.
Simulation and real-world experiments demonstrate the effectiveness and practicality of the framework. VLM-UDMC represents a significant step forward in VLM-augmented autonomous driving technology, offering a robust and adaptable solution for modern urban mobility. Future research will focus on addressing remaining challenges, including accurate spatial understanding, multi-sized camera image compatibility, and lifelong learning, to enhance performance and expand the framework’s applicability, bringing fully autonomous urban driving closer to realization.

\bibliographystyle{ieeetr}
\bibliography{refs}

\begin{thebibliography}{10}

\bibitem{chen2022milestones}
L.~Chen, Y.~Li, C.~Huang, B.~Li, Y.~Xing, D.~Tian, L.~Li, Z.~Hu, X.~Na, Z.~Li, {\em et~al.}, ``Milestones in autonomous driving and intelligent vehicles: Survey of surveys,'' {\em IEEE Transactions on Intelligent Vehicles}, vol.~8, no.~2, pp.~1046--1056, 2022.

\bibitem{gao2022optimal}
Z.~Gao, Z.~Wu, W.~Hao, K.~Long, Y.-J. Byon, and K.~Long, ``Optimal trajectory planning of connected and automated vehicles at on-ramp merging area,'' {\em IEEE Transactions on Intelligent Transportation Systems}, vol.~23, no.~8, pp.~12675--12687, 2022.

\bibitem{zhang2023adaptive}
Y.~Zhang, X.~Liang, D.~Li, S.~S. Ge, B.~Gao, H.~Chen, and T.~H. Lee, ``Adaptive safe reinforcement learning with full-state constraints and constrained adaptation for autonomous vehicles,'' {\em IEEE Transactions on Cybernetics}, vol.~54, no.~3, pp.~1907--1920, 2023.

\bibitem{gao2024enhance}
Z.~Gao, Y.~Mu, C.~Chen, J.~Duan, P.~Luo, Y.~Lu, and S.~Eben~Li, ``Enhance sample efficiency and robustness of {End-to-End} urban autonomous driving via semantic masked world model,'' {\em IEEE Transactions on Intelligent Transportation Systems}, vol.~25, no.~10, pp.~13067--13079, 2024.

\bibitem{zhao2024enhanced}
Y.~Zhao, L.~Wang, X.~Yun, C.~Chai, Z.~Liu, W.~Fan, X.~Luo, Y.~Liu, and X.~Qu, ``Enhanced scene understanding and situation awareness for autonomous vehicles based on semantic segmentation,'' {\em IEEE Transactions on Systems, Man, and Cybernetics: Systems}, vol.~54, no.~11, pp.~6537--6549, 2024.

\bibitem{jia2023interactive}
S.~Jia, Y.~Zhang, X.~Li, X.~Na, Y.~Wang, B.~Gao, B.~Zhu, and R.~Yu, ``Interactive decision-making with switchable game modes for automated vehicles at intersections,'' {\em IEEE Transactions on Intelligent Transportation Systems}, vol.~24, no.~11, pp.~11785--11799, 2023.

\bibitem{fan2018baidu}
H.~Fan, F.~Zhu, C.~Liu, L.~Zhang, L.~Zhuang, D.~Li, W.~Zhu, J.~Hu, H.~Li, and Q.~Kong, ``Baidu apollo {EM} motion planner,'' {\em arXiv preprint arXiv:1807.08048}, 2018.

\bibitem{jiang2023vad}
B.~Jiang, S.~Chen, Q.~Xu, B.~Liao, J.~Chen, H.~Zhou, Q.~Zhang, W.~Liu, C.~Huang, and X.~Wang, ``{VAD}: Vectorized scene representation for efficient autonomous driving,'' in {\em Proceedings of the IEEE/CVF International Conference on Computer Vision}, pp.~8340--8350, 2023.

\bibitem{chen2023milestones2}
L.~Chen, Y.~Li, C.~Huang, Y.~Xing, D.~Tian, L.~Li, Z.~Hu, S.~Teng, C.~Lv, J.~Wang, D.~Cao, N.~Zheng, and F.-Y. Wang, ``Milestones in autonomous driving and intelligent vehicles—{Part I}: Control, computing system design, communication, {HD} map, testing, and human behaviors,'' {\em IEEE Transactions on Systems, Man, and Cybernetics: Systems}, vol.~53, no.~9, pp.~5831--5847, 2023.

\bibitem{zhang2017finite}
M.~Zhang, N.~Li, A.~Girard, and I.~Kolmanovsky, ``A finite state machine based automated driving controller and its stochastic optimization,'' in {\em Dynamic Systems and Control Conference}, vol.~58288, p.~V002T07A002, American Society of Mechanical Engineers, 2017.

\bibitem{qi2022hierarchical}
Y.~Qi, B.~He, R.~Wang, L.~Wang, and Y.~Xu, ``Hierarchical motion planning for autonomous vehicles in unstructured dynamic environments,'' {\em IEEE Robotics and Automation Letters}, vol.~8, no.~2, pp.~496--503, 2022.

\bibitem{dosovitskiy2017carla}
A.~Dosovitskiy, G.~Ros, F.~Codevilla, A.~Lopez, and V.~Koltun, ``{CARLA}: An open urban driving simulator,'' in {\em Conference on Robot Learning}, pp.~1--16, PMLR, 2017.

\bibitem{pan2024safe}
H.~Pan, M.~Luo, J.~Wang, T.~Huang, and W.~Sun, ``A safe motion planning and reliable control framework for autonomous vehicles,'' {\em IEEE Transactions on Intelligent Vehicles}, vol.~9, no.~4, pp.~4780--4793, 2024.

\bibitem{xu2019design}
S.~Xu and H.~Peng, ``Design, analysis, and experiments of preview path tracking control for autonomous vehicles,'' {\em IEEE Transactions on Intelligent Transportation Systems}, vol.~21, no.~1, pp.~48--58, 2019.

\bibitem{brudigam2021stochastic}
T.~Br{\"u}digam, M.~Olbrich, D.~Wollherr, and M.~Leibold, ``Stochastic model predictive control with a safety guarantee for automated driving,'' {\em IEEE Transactions on Intelligent Vehicles}, vol.~8, no.~1, pp.~22--36, 2021.

\bibitem{liu2025udmc}
H.~Liu, K.~Chen, Y.~Li, Z.~Huang, M.~Liu, and J.~Ma, ``{UDMC}: Unified decision-making and control framework for urban autonomous driving with motion prediction of traffic participants,'' {\em IEEE Transactions on Intelligent Transportation Systems}, vol.~26, no.~5, pp.~5856--5871, 2025.

\bibitem{hu2023planning}
Y.~Hu, J.~Yang, L.~Chen, K.~Li, C.~Sima, X.~Zhu, S.~Chai, S.~Du, T.~Lin, W.~Wang, {\em et~al.}, ``Planning-oriented autonomous driving,'' in {\em Proceedings of the IEEE/CVF Conference on Computer Vision and Pattern Recognition}, pp.~17853--17862, 2023.

\bibitem{liu2025vlm}
P.~Liu, H.~Liu, H.~Liu, X.~Liu, J.~Ni, and J.~Ma, ``{VLM-E2E}: Enhancing end-to-end autonomous driving with multimodal driver attention fusion,'' {\em arXiv preprint arXiv:2502.18042}, 2025.

\bibitem{liu2025dsdrive}
W.~Liu, P.~Liu, and J.~Ma, ``{DSDrive}: Distilling large language model for lightweight end-to-end autonomous driving with unified reasoning and planning,'' {\em arXiv preprint arXiv:2505.05360}, 2025.

\bibitem{liao2025diffusiondrive}
B.~Liao, S.~Chen, H.~Yin, B.~Jiang, C.~Wang, S.~Yan, X.~Zhang, X.~Li, Y.~Zhang, Q.~Zhang, {\em et~al.}, ``{DiffusionDrive}: Truncated diffusion model for end-to-end autonomous driving,'' in {\em Proceedings of the Computer Vision and Pattern Recognition Conference}, pp.~12037--12047, 2025.

\bibitem{zhou2025opendrivevla}
X.~Zhou, X.~Han, F.~Yang, Y.~Ma, and A.~C. Knoll, ``{OpenDriveVLA}: Towards end-to-end autonomous driving with large vision language action model,'' {\em arXiv preprint arXiv:2503.23463}, 2025.

\bibitem{li2024think}
Q.~Li, X.~Jia, S.~Wang, and J.~Yan, ``{Think2Drive}: Efficient reinforcement learning by thinking in latent world model for quasi-realistic autonomous driving (in {CARLA}-v2),'' in {\em European Conference on Computer Vision}, 2024.

\bibitem{kuznietsov2024explainable}
A.~Kuznietsov, B.~Gyevnar, C.~Wang, S.~Peters, and S.~V. Albrecht, ``Explainable ai for safe and trustworthy autonomous driving: A systematic review,'' {\em IEEE Transactions on Intelligent Transportation Systems}, vol.~25, no.~12, pp.~19342--19364, 2024.

\bibitem{long2024vlm}
K.~Long, H.~Shi, J.~Liu, and X.~Li, ``{VLM-MPC}: {Vision Language Foundation Model}-guided model predictive controller for autonomous driving,'' {\em arXiv preprint arXiv:2408.04821}, 2024.

\bibitem{atsuta2025lvlm}
K.~Atsuta, K.~Honda, H.~Okuda, and T.~Suzuki, ``{LVLM-MPC}: Collaboration for autonomous driving: A safety-aware and task-scalable control architecture,'' {\em arXiv preprint arXiv:2505.04980}, 2025.

\bibitem{yao2024calmm}
R.~Yao, Y.~Wang, H.~Liu, R.~Yang, Z.~Peng, L.~Zhu, and J.~Ma, ``{CALMM-Drive:} confidence-aware autonomous driving with large multimodal model,'' {\em arXiv preprint arXiv:2412.04209}, 2024.

\bibitem{lstm_ts_ctcs}
H.~Guo, H.~Zhu, J.~Wang, V.~Prahlad, W.~K. Ho, C.~W. de~Silva, and T.~H. Lee, ``Lightweight compressed temporal and compressed spatial attention with augmentation fusion in remaining useful life prediction,'' in {\em 49th Annual Conference of the IEEE Industrial Electronics Society}, pp.~1--6, 2023.

\bibitem{wen2023transformerstimeseriessurvey}
Q.~Wen, T.~Zhou, C.~Zhang, W.~Chen, Z.~Ma, J.~Yan, and L.~Sun, ``Transformers in time series: a survey,'' in {\em Proceedings of the International Joint Conference on Artificial Intelligence}, pp.~6778--6786, 2023.

\bibitem{tian2024drivevlm}
X.~Tian, J.~Gu, B.~Li, Y.~Liu, Y.~Wang, Z.~Zhao, K.~Zhan, P.~Jia, X.~Lang, and H.~Zhao, ``Drive{VLM}: The convergence of autonomous driving and large vision-language models,'' in {\em 8th Annual Conference on Robot Learning}, 2024.

\bibitem{li2024dcoma}
L.~Li, C.~Qian, J.~Gan, D.~Zhang, X.~Qu, F.~Xiao, and B.~Ran, ``{DCoMA}: A dynamic coordinative merging assistant strategy for on-ramp vehicles with mixed traffic conditions,'' {\em Transportation Research Part C: Emerging Technologies}, vol.~165, p.~104700, 2024.

\bibitem{liu2024improved}
H.~Liu, Z.~Huang, Z.~Zhu, Y.~Li, S.~Shen, and J.~Ma, ``Improved consensus admm for cooperative motion planning of large-scale connected autonomous vehicles with limited communication,'' {\em IEEE Transactions on Intelligent Vehicles}, 2024.

\bibitem{liu2025codrivevlm}
H.~Liu, R.~Yao, W.~Liu, Z.~Huang, S.~Shen, and J.~Ma, ``{CoDriveVLM}: {VLM}-enhanced urban cooperative dispatching and motion planning for future autonomous mobility on demand systems,'' {\em arXiv preprint arXiv:2501.06132}, 2025.

\bibitem{lu2025empowering}
H.~Lu, M.~Zhu, C.~Lu, S.~Feng, X.~Wang, Y.~Wang, and H.~Yang, ``Empowering safer socially sensitive autonomous vehicles using human-plausible cognitive encoding,'' {\em Proceedings of the National Academy of Sciences}, vol.~122, no.~21, p.~e2401626122, 2025.

\bibitem{liu2024incremental}
H.~Liu, K.~Chen, and J.~Ma, ``Incremental learning-based real-time trajectory prediction for autonomous driving via sparse {Gaussian} process regression,'' in {\em 2024 IEEE Intelligent Vehicles Symposium}, pp.~1--7, 2024.

\bibitem{zeng2023transformers}
A.~Zeng, M.~Chen, L.~Zhang, and Q.~Xu, ``Are {Transformers} effective for time series forecasting?,'' in {\em Proceedings of the AAAI conference on artificial intelligence}, pp.~11121--11128, 2023.

\bibitem{ge2021numerically}
Q.~Ge, Q.~Sun, S.~E. Li, S.~Zheng, W.~Wu, and X.~Chen, ``Numerically stable dynamic bicycle model for discrete-time control,'' in {\em IEEE Intelligent Vehicles Symposium}, pp.~128--134, 2021.

\bibitem{marafioti2025smolvlm}
A.~Marafioti, O.~Zohar, M.~Farré, M.~Noyan, E.~Bakouch, P.~Cuenca, C.~Zakka, L.~B. Allal, A.~Lozhkov, N.~Tazi, V.~Srivastav, J.~Lochner, H.~Larcher, M.~Morlon, L.~Tunstall, L.~von Werra, and T.~Wolf, ``{SmolVLM}: Redefining small and efficient multimodal models,'' {\em arXiv preprint arXiv:2504.05299}, 2025.

\bibitem{qwen3technicalreport}
A.~Yang, A.~Li, B.~Yang, B.~Zhang, B.~Hui, B.~Zheng, B.~Yu, C.~Gao, C.~Huang, C.~Lv, {\em et~al.}, ``Qwen3 technical report,'' {\em arXiv preprint arXiv:2505.09388}, 2025.

\bibitem{li2025drive}
Y.~Li, M.~Tian, D.~Zhu, J.~Zhu, Z.~Lin, Z.~Xiong, and X.~Zhao, ``{Drive-R1}: Bridging reasoning and planning in {VLMs} for autonomous driving with reinforcement learning,'' {\em arXiv preprint arXiv:2506.18234}, 2025.

\bibitem{radford2021learning}
A.~Radford, J.~W. Kim, C.~Hallacy, A.~Ramesh, G.~Goh, S.~Agarwal, G.~Sastry, A.~Askell, P.~Mishkin, J.~Clark, {\em et~al.}, ``Learning transferable visual models from natural language supervision,'' in {\em International Conference on Machine Learning}, pp.~8748--8763, 2021.

\bibitem{shao2023safety}
H.~Shao, L.~Wang, R.~Chen, H.~Li, and Y.~Liu, ``Safety-enhanced autonomous driving using interpretable sensor fusion transformer,'' in {\em Conference on Robot Learning}, pp.~726--737, 2023.

\bibitem{shalev2017formal}
S.~Shalev-Shwartz, S.~Shammah, and A.~Shashua, ``On a formal model of safe and scalable self-driving cars,'' {\em arXiv preprint arXiv:1708.06374}, 2017.

\bibitem{Andersson2019}
J.~A.~E. Andersson, J.~Gillis, G.~Horn, J.~B. Rawlings, and M.~Diehl, ``{CasADi} -- {A} software framework for nonlinear optimization and optimal control,'' {\em Mathematical Programming Computation}, vol.~11, no.~1, pp.~1--36, 2019.

\end{thebibliography}
\end{document}